\documentclass{article} 
\usepackage{iclr2025_conference,times}
\def\confDataName{Swift4D}

\usepackage{amsmath,amsfonts,bm}

\def\eqref#1{equation~\ref{#1}}

\def\1{\bm{1}}

\DeclareMathAlphabet{\mathsfit}{\encodingdefault}{\sfdefault}{m}{sl}
\SetMathAlphabet{\mathsfit}{bold}{\encodingdefault}{\sfdefault}{bx}{n}

\usepackage{hyperref}
\usepackage{url}
\usepackage{multicol} 
\usepackage{array} 
\usepackage{threeparttable}
\usepackage{cutwin}
\usepackage{subcaption}
\usepackage{graphicx} 
\usepackage{wrapfig}
\usepackage{xcolor}
\usepackage{colortbl}
\usepackage[none]{hyphenat} 
\usepackage{booktabs}
\usepackage{multirow}
\usepackage{dsfont}

\title{\confDataName: Adaptive divide-and-conquer Gaussian Splatting for compact and efficient reconstruction of dynamic scene}

\author{%
  Jiahao Wu, \quad Rui Peng, \quad Zhiyan Wang, \quad Lu Xiao,  \\
  \textbf{Luyang Tang}, \quad \textbf{Jinbo Yan}, \quad \textbf{Kaiqiang Xiong},  \quad \textbf{Ronggang Wang}\thanks{Corresponding author }\\
  Guangdong Provincial Key Laboratory of Ultra High Definition Immersive Media Technology \\
  Shenzhen Graduate School, Peking University\\
  \texttt{2301212750@stu.pku.edu.cn}, \quad \texttt{rgwang@pkusz.edu.cn} 
}

\iclrfinalcopy
\begin{document}

\maketitle

\vspace{-3.3em}
\begin{figure}[htbp]
    \centering
    \begin{minipage}{0.5\textwidth}
        \centering
        \begin{minipage}{0.49\textwidth}
            \centering
            \includegraphics[width=\textwidth]{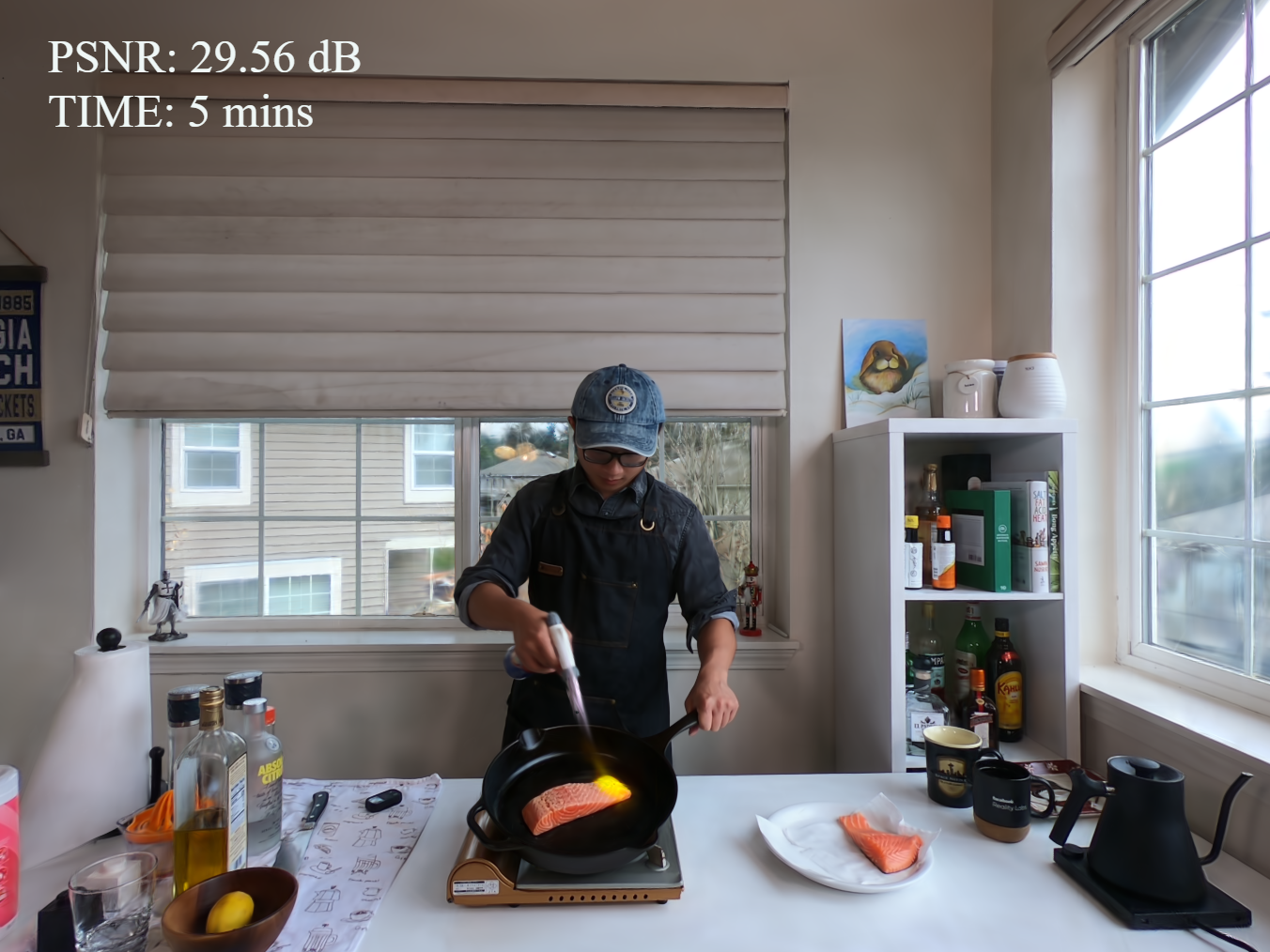}
        \end{minipage}
        \hfill
        \begin{minipage}{0.49\textwidth}
            \centering
            \includegraphics[width=\textwidth]{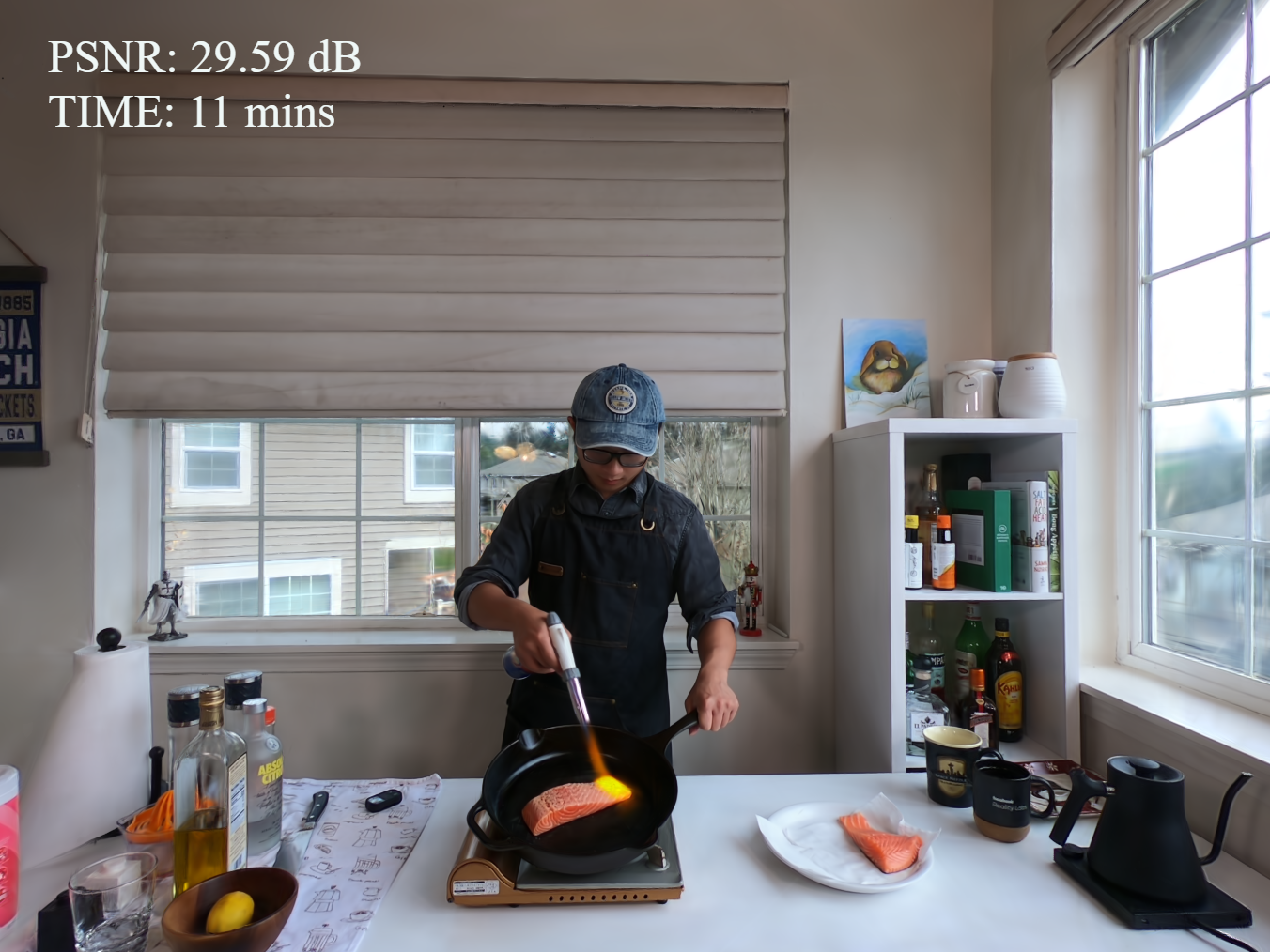}
        \end{minipage}
        \vfill
        \begin{minipage}{0.49\textwidth}
            \centering
            \includegraphics[width=\textwidth]{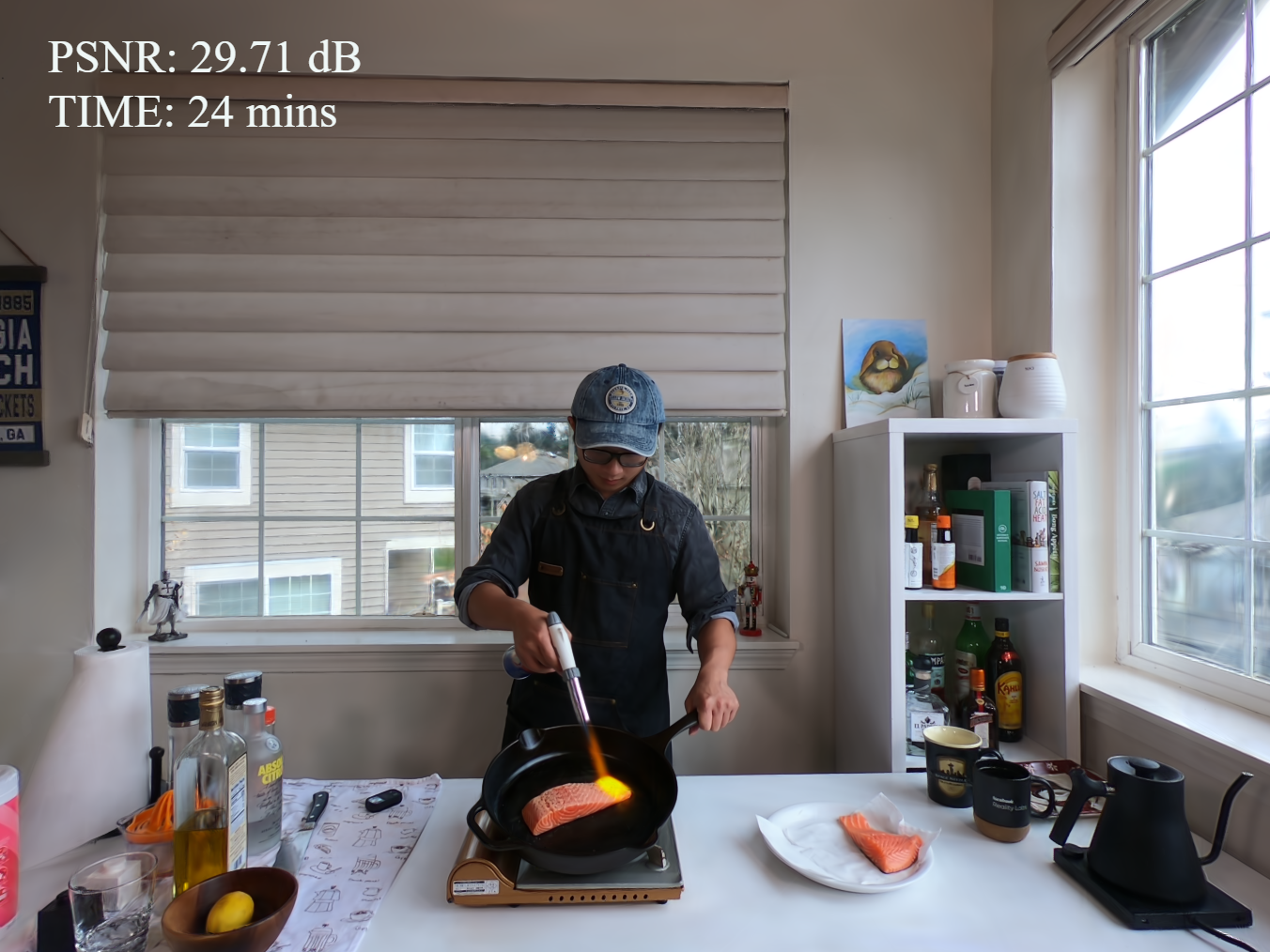}
        \end{minipage}
        \hfill
        \begin{minipage}{0.49\textwidth}
            \centering
            \includegraphics[width=\textwidth]{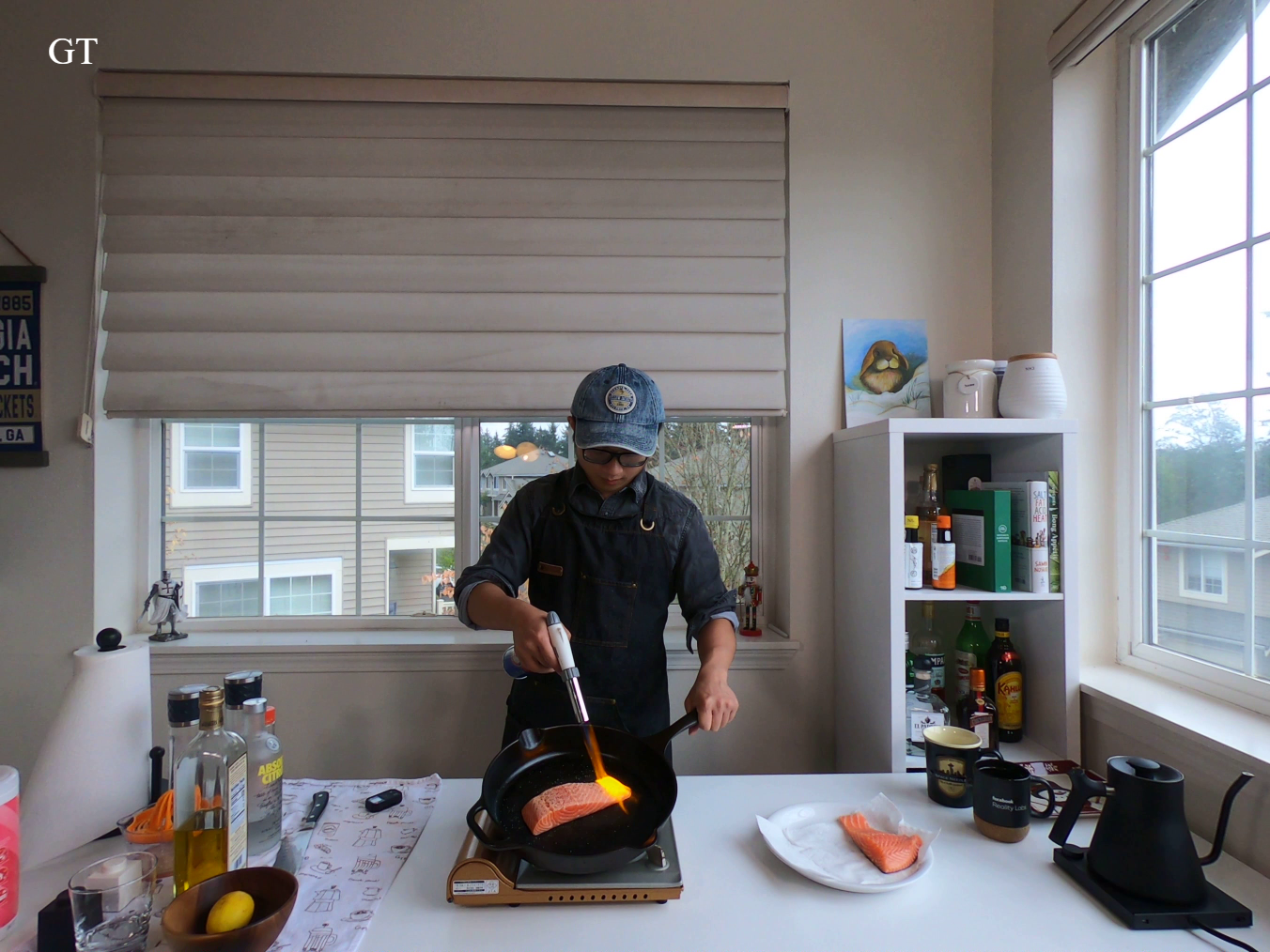}
        \end{minipage}
    \end{minipage}
    \hfill
    \begin{minipage}{0.40\textwidth}
        \centering
        \includegraphics[width=\textwidth, trim=3.6cm 2cm 13cm 0cm]{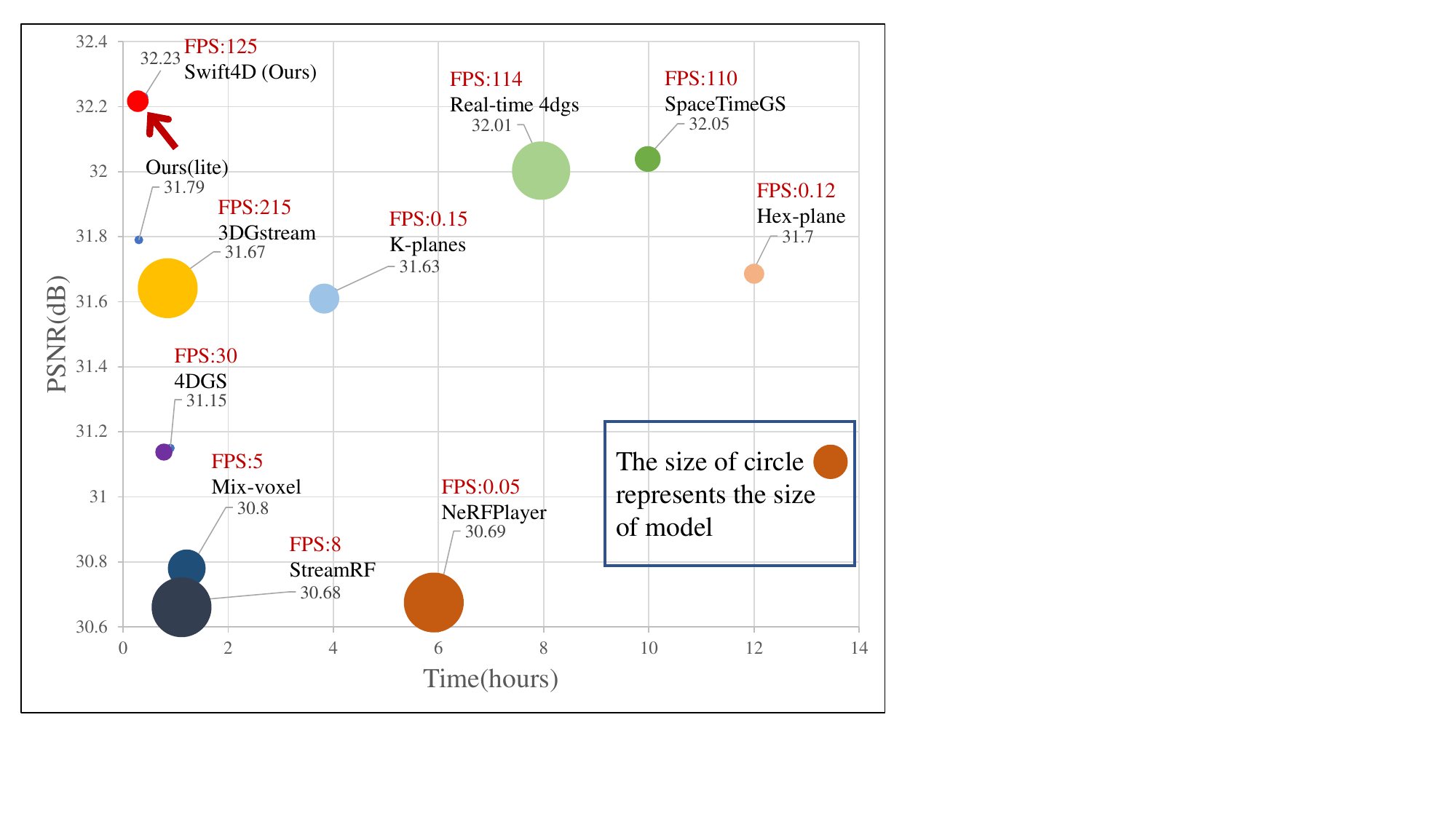}
    \end{minipage}
    \caption{Our method demonstrates high-quality rendering, rapid convergence, and compact storage characteristics. It can achieve competitive result with just 5 minutes of training. Additionally, with increased training iterations, our method excels in handling finer details.}
    \label{figs:dsseg}
\end{figure}

\begin{abstract}

Novel view synthesis has long been a practical but challenging task, although the introduction of numerous methods to solve this problem, even combining advanced representations like 3D Gaussian Splatting, they still struggle to recover high-quality results and often consume too much storage memory and training time. 
In this paper we propose Swift4D, a divide-and-conquer 3D Gaussian Splatting method that can handle static and dynamic primitives separately, achieving a good trade-off between rendering quality and efficiency, motivated by the fact that most of the scene is the static primitive and does not require additional dynamic properties. Concretely, we focus on modeling dynamic transformations only for the dynamic primitives which benefits both efficiency and quality. We first employ a learnable decomposition strategy to separate the primitives, which relies on an additional parameter to classify primitives as static or dynamic. For the dynamic primitives, we employ a compact multi-resolution 4D Hash mapper to transform these primitives from canonical space into deformation space at each timestamp, and then mix the static and dynamic primitives to produce the final output. This divide-and-conquer method facilitates efficient training and reduces storage redundancy. Our method not only achieves state-of-the-art rendering quality while being 20× faster in training than previous SOTA methods with a minimum storage requirement of only 30MB on real-world datasets. Code is available at \url{https://github.com/WuJH2001/swift4d}.


\end{abstract}
\section{Introduction}

\begin{figure}[t]
    \centering
    \includegraphics[width= \textwidth ,trim=1cm 11cm 2cm 3.3cm]{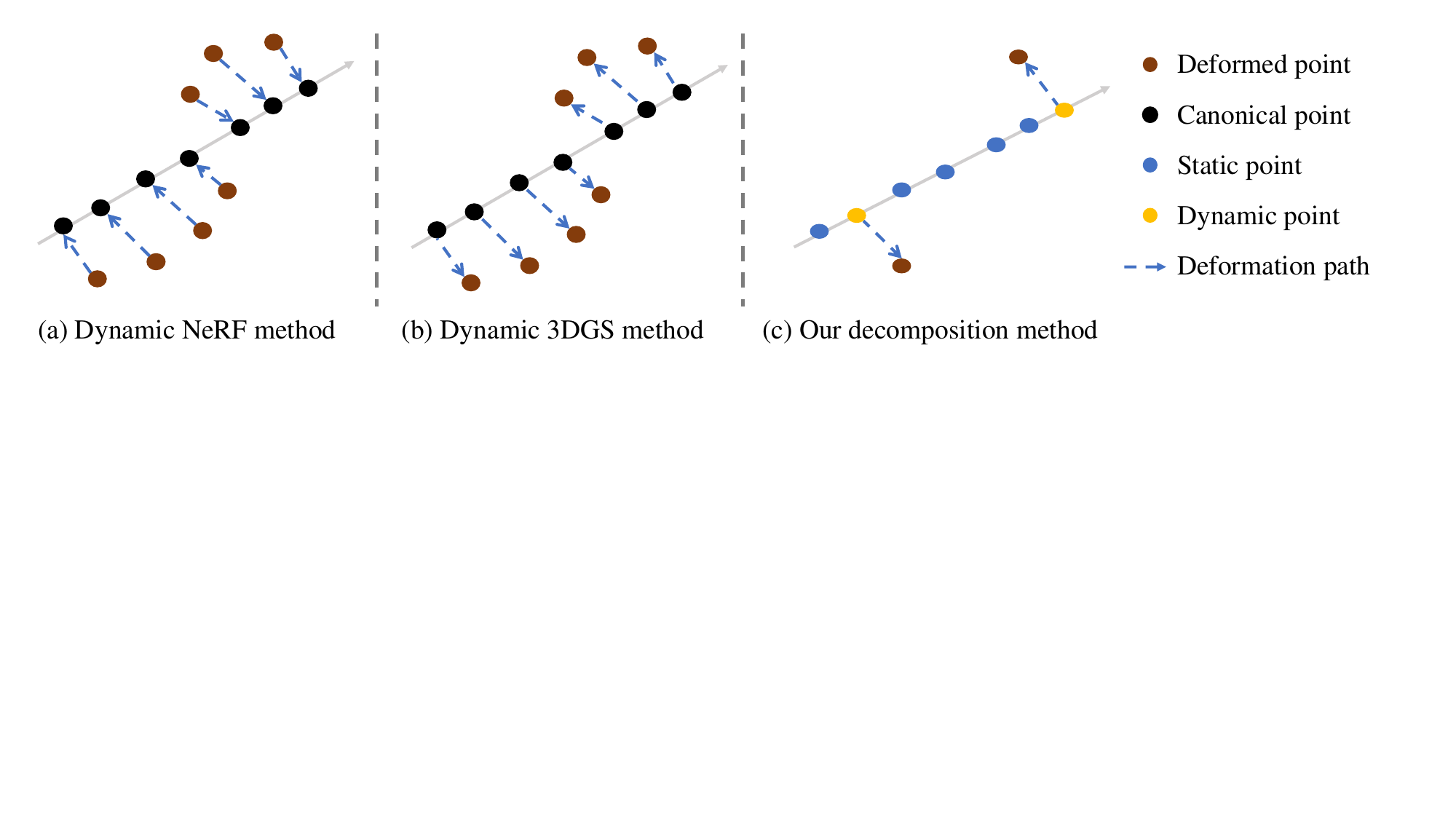}
    \caption{ \textbf{Illustration of different dynamic scene rendering methods.} (a) \cite{pumarola2021d,park2021nerfies} proposes mapping deformation field points to canonical space, a widely adopted practice in NeRF-based methods; (b) \cite{wu20244d, yang2024deformable} propose mapping canonical space points to the deformation field; (c) We propose dividing the points in canonical space into dynamic and static, and then mapping only the dynamic points to the deformation space.}
    \label{figs:decomposition demo}
    \vspace{-1.5em}
\end{figure}

Novel view synthesis (NVS) is a crucial task in computer vision and graphics, with significant applications in areas such as augmented reality (AR), virtual reality (VR), and content production. The goal of NVS is to render photorealistic images from arbitrary viewpoints using 2D images or video inputs. While recent advancements have achieved considerable success in static scenes, this task becomes particularly challenging when applied to dynamic scenes, where complexities introduced by object motion and temporal changes make accurate rendering significantly difficult.


Current NVS techniques can be broadly classified into two predominant approaches: neural rendering methods, exemplified by Neural Radiance Fields (NeRF) \cite{mildenhall2021nerf}, and point cloud-based rendering techniques, such as 3D Gaussian Splatting \cite{kerbl20233d}.
NeRFs have recently made significant strides in achieving photorealistic rendering of static scenes, with subsequent works \cite{barron2021mip,barron2022mip,barron2023zip,reiser2023merf} further enhancing both quality and speed. Despite these advancements in static scene rendering, NeRFs face significant challenges when extended to dynamic scenes, primarily due to the substantial training time and storage requirements. To overcome these obstacles, various approaches have been proposed. As shown in Fig.\ref{figs:decomposition demo}(a),  \cite{pumarola2021d} and \cite{park2021nerfies} leverage deformation fields to map deformation space at arbitrary timestamps to canonical space, effectively capturing dynamic scene changes. \cite{li2021neural} and \cite{gao2021dynamic} employ scene flow to model the motion trajectories within dynamic environments.  \cite{cao2023hexplane} and \cite{fridovich2023k}, decompose the 4D spacetime domain into multiple compact planes, thereby improving training and rendering speeds. Although these methods have achieved some degree of success, achieving high-quality real-time rendering remains challenging.


Compared to NeRF, 3DGS offers significant advantages, including real-time rendering and substantially reduced training time. Within the scope of dynamic modeling, several notable methods have emerged. As shown in Fig.\ref{figs:decomposition demo}(b), 4DGS \cite{wu20244d}, inspired by HexPlane, introduces a neural voxel encoder to model deformation relationships over time. 
3DGStream \cite{sun20243dgstream} utilizes a compact Neural Transformation Cache (NTC) to efficiently model the translation and rotation of 3D Gaussians between two adjacent frames.
RTGS \cite{yang2023real} treats spacetime as an integrated whole by optimizing a set of 4D primitives, parameterized as anisotropic ellipses that capture both geometry and appearance.
Additionally, STGS \cite{li2024spacetime} enhances standard 3D Gaussians with temporal opacity and motion/rotation parameters, effectively capturing both static and dynamic elements to model dynamic deformation.

While these approaches achieve higher-quality results with faster rendering times, they still face challenges related to long training time and heavy storage requirements.
One potential limitation in their approach is the uniform treatment of all Gaussian points during the modeling process. However, we observe that static points, such as those in background regions, constitute the majority of the scene. These points exhibit minimal or no deformation and therefore do not require complex dynamic modeling. It is more efficient to partition the scene into static and dynamic points and model each separately. This strategy has the potential to significantly reduce computational overhead and storage requirements. Moreover, as demonstrated by \cite{wang2023mixed}, applying the same modeling technique to both dynamic and static points can cause blurring in dynamic regions due to the influence of static areas, ultimately compromising rendering quality.

In this paper, we introduce \confDataName, a method that simultaneously achieves fast convergence, compact storage, and real-time high-quality rendering. Our approach starts by decomposing Gaussian points into dynamic and static groups based on 2D multi-view images, incorporating an additional parameter \(d\) for differentiation.
For temporal modeling, we employ a deformation field approach using a compact multi-resolution 4DHash and MLPs as the deformer, which maps dynamic Gaussian points from canonical space to deformation space at arbitrary timestamps. Notably, as shown in Fig. \ref{figs:decomposition demo}(c), temporal modeling is applied exclusively to the dynamic points, while static points are treated as temporally invariant, significantly reducing computational demands. This reduction in the number of dynamic points enables the 4DHash to concentrate on deformation, leading to faster convergence and improved rendering quality. Finally, we combine the static and dynamic Gaussian points to render the final output. This approach also addresses the issue of blurring caused by static elements interfering with the time-aware multi-resolution 4DHash.

Our method achieves SOTA performance in terms of training and rendering speed, storage efficiency, and rendering quality. Furthermore,  our supplement videos ( basketball 1 and 2) demonstrate that our approach remains effective even in scenarios involving large movements. We will release our code and pre-trained models upon acceptance.
In summary, the key contributions of our work are:
\begin{enumerate}
\item [1)] We propose a novel method for decomposing dynamic 3D scenes into dynamic and static components based on 2D images, effectively reducing computational complexity. This method can be seamlessly integrated into existing dynamic approaches as a plug-and-play module to enhance quality.
\item[2)] We introduce a compact multi-resolution 4DHash, with a footprint as small as 8MB, to effectively model the spatio-temporal domain. This approach not only enhances rendering quality and accelerates training but also ensures efficient and compact storage.
\item[3)] Our method achieves state-of-the-art performance in training and rendering speed, storage, and high-quality output.

\end{enumerate}

\section{Related work}
\label{sec: Related work}
\textbf{Novel View Synthesis.} 
In recent years, novel view synthesis has garnered significant attention, leading to numerous breakthroughs. NeRF\cite{mildenhall2021nerf} pioneered this domain by leveraging multi-layer perceptrons (MLPs) combined with volume rendering to model 3D radiance fields, enabling image rendering from arbitrary viewpoints. Subsequent works aimed to enhance efficiency and quality. Methods such as TensorF \cite{chen2022tensorf}, DVGO \cite{sun2022direct}, Plenoxel\cite{fridovich2022plenoxels}, and Plenoctree \cite{yu2021plenoctrees} adopt grid-based representations for faster training and rendering. Instant NGP \cite{muller2022instant} further accelerates this process with a hash encoder, significantly reducing computation time. Meanwhile, MipNeRF \cite{barron2021mip} and MipNeRF360 \cite{barron2022mip} propose integrated positional encoding (IPE) to model conical frustums, effectively mitigating aliasing issues.
More recently, 3DGS \cite{kerbl20233d} introduced a novel point-based rendering paradigm for novel view synthesis, achieving real-time rendering with high quality. This has spurred additional advancements, including Mipsplatting \cite{yu2024mip} for anti-aliasing, 2DGS \cite{huang20242d} for improved mesh extraction, and ScaffoldGS \cite{lu2024scaffold} for large-scale scene rendering.

\begin{figure}[t]
    \centering
    \includegraphics[width= \textwidth ,trim=0cm 4cm 2.2cm 3.3cm]{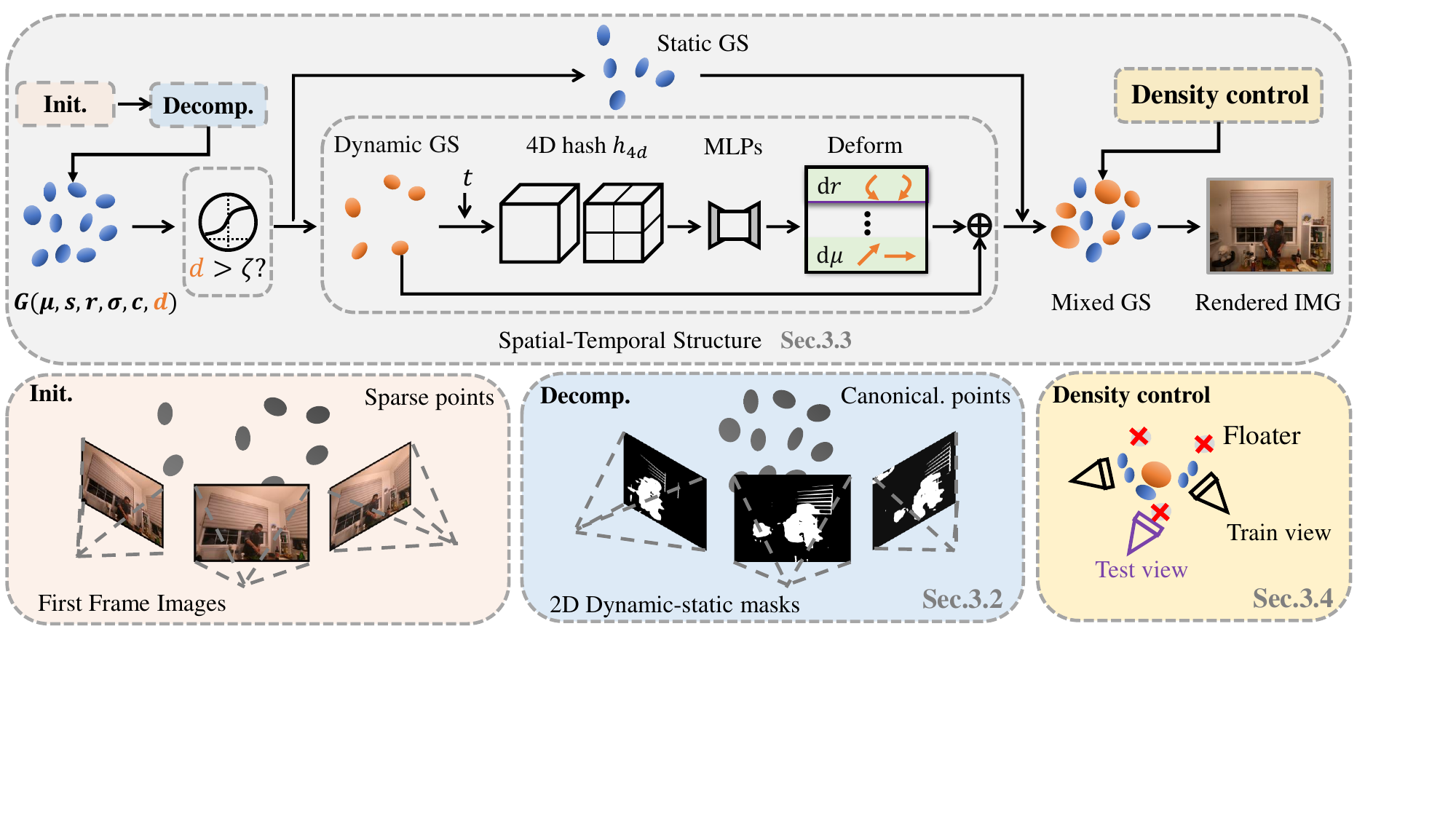}
    \caption{\textbf{ Pipeline of our \confDataName.} First, we use the first frame images to obtain a well-initialized canonical point cloud. Then, we train the dynamic parameter $d$ according to the method described in Sec.\ref{sec:segmentation}. Based on $d$, the point cloud is divided into dynamic and static categories. Dynamic points undergo deformation using a spatio-temporal structure, as discussed in Sec.\ref{sec:4dhash}. Finally, the deformed dynamic points are mixed with static points for rendering.}
    \label{figs:pipelins}
\end{figure}

\textbf{Novel View Synthesis for dynamic scene.}  \cite{li2021neural, lin2024gaussian, kratimenos2023dynmf} attempt to directly model the trajectories of moving points across the scene, but they continue to encounter challenges related to storage. \cite{pumarola2021d, park2021nerfies, wu20244d, yang2024deformable}
try to build a consistent canonical space across each time step and then employ a deformer, mainly MLP-based and Muti-plane-based,  to map this canonical space to deformation spaces at each timestamp. 
\cite{huang2024sc} focuses on monocular dynamic inputs, leveraging sparse control points to reconstruct scene dynamics with exceptionally high FPS.
\cite{lin2024gaussian} employs Fourier series and polynomial fitting to model the motion of Gaussian points, enabling dynamic reconstruction. 
K-planes \cite{fridovich2023k} and Hexplane \cite{cao2023hexplane} employ an explicit structural representation of the 6D light field rather than modeling underlying motions.
Representing the deformer using MLPs or low-rank planes can reduce storage requirements, but it often results in slower training and limited capacity for capturing complex deformations.

Recently, \cite{he2024s4d,yan20244d} attempt to separate dynamic and static Gaussian points to improve rendering quality and introduced external models to segment foreground and background areas. While these efforts have explored this direction, the resulting output quality remains suboptimal.  \cite{liang2023gaufre} employs adaptive dynamic-static separation, which differs from our explicit separation approach.


\section{Method}
\label{sec:method}


Our main approach aims to achieve faster training speeds and higher quality rendering results through the decomposition of dynamic and static elements. Based on this insight, we designed our pipeline, as illustrated in Fig. \ref{figs:pipelins}. In this section, we will provide a detailed analysis of each module in the pipeline.
The preliminary concepts of 3D Gaussian Splatting are briefly introduced in Sec. \ref{sec:preliminary}. We initially train the canonical space Gaussians using the first-frame images and then optimize the dynamic parameter \(d\) of each Gaussian point based on the 2D dynamic-static pixel masks from different viewpoints, as discussed in Sec.\ref{sec:segmentation}. In the following stage, as outlined in Sec. \ref{sec:4dhash}, we freeze the training of dynamic parameter \(d\) and proceed to jointly optimize the remaining parameter of the Gaussian points alongside the swift spatio-temporal structure. Furthermore, our pruning strategies are thoroughly described in Sec. \ref{sec:densification and pruning}, while Sec. \ref{sec:optimization} provides an in-depth discussion of the optimization process.

\subsection{Gaussian Splatting Preliminary}
\label{sec:preliminary}
3DGS\cite{kerbl20233d} uses 3D Gaussian points as its rendering primitives. These 3D Gaussian points have the following parameter: mean \(\mu\), covariance matrix \(\Sigma\) , opacity \(\sigma\) , and view-dependent color \(c\). A 3D Gaussian point is mathematically defined as: 
\begin{equation}
 G(x) = e^{-\frac{1}{2}(x-\mu)^T\Sigma^{-1}(x-\mu)} 
\end{equation}
In the next rendering phase, the 3D mean  \(\mu\) is directly projected onto the plane as a 2D mean \(\mu^{2D}\), while the 3D covariance matrix is transformed into a 2D covariance matrix using the following formula: \(\Sigma' = (JW\Sigma W^T J^T)\), where \(W\) and \(J\) denote the viewing transformation and the Jacobian of the affine approximation of the perspective projection transformation, respectively. Finally, the color of each pixel can be calculated using the following formula: 

\begin{equation}
\label{eq:blending}
C(x)=\sum_{i\in\mathcal{N(\mathbf{x})}}\mathbf{c}_i\alpha_i(x)\prod_{j=1}^{i-1}(1-\alpha_j(x)) ~
\text{where}~
    \alpha_i(x)=\sigma_i\exp\left(-\frac{1}{2}(x-\mu^{2D}_i)^T\Sigma'^{-1}(x-\mu^{2D}_i)\right).
\end{equation}
Where \(N\) is the number of Gaussian points that intersect with the pixel $x \in \mathds{R}^2$.   In the actual implementation, the covariance matrix \(\Sigma\) is typically decomposed into rotation \(q\) and scaling \(s\). The color \(c\) is represented by a spherical harmonics (SH) function. Therefore, a Gaussian point can be represented as \(G\{\mu, q, s, \sigma, c\}\).

\begin{figure}[t]
\vspace{-1em}
    \centering
    \begin{minipage}{0.40\textwidth}
        \centering
        \includegraphics[width=\textwidth, trim=6cm 5cm 12cm 3cm]{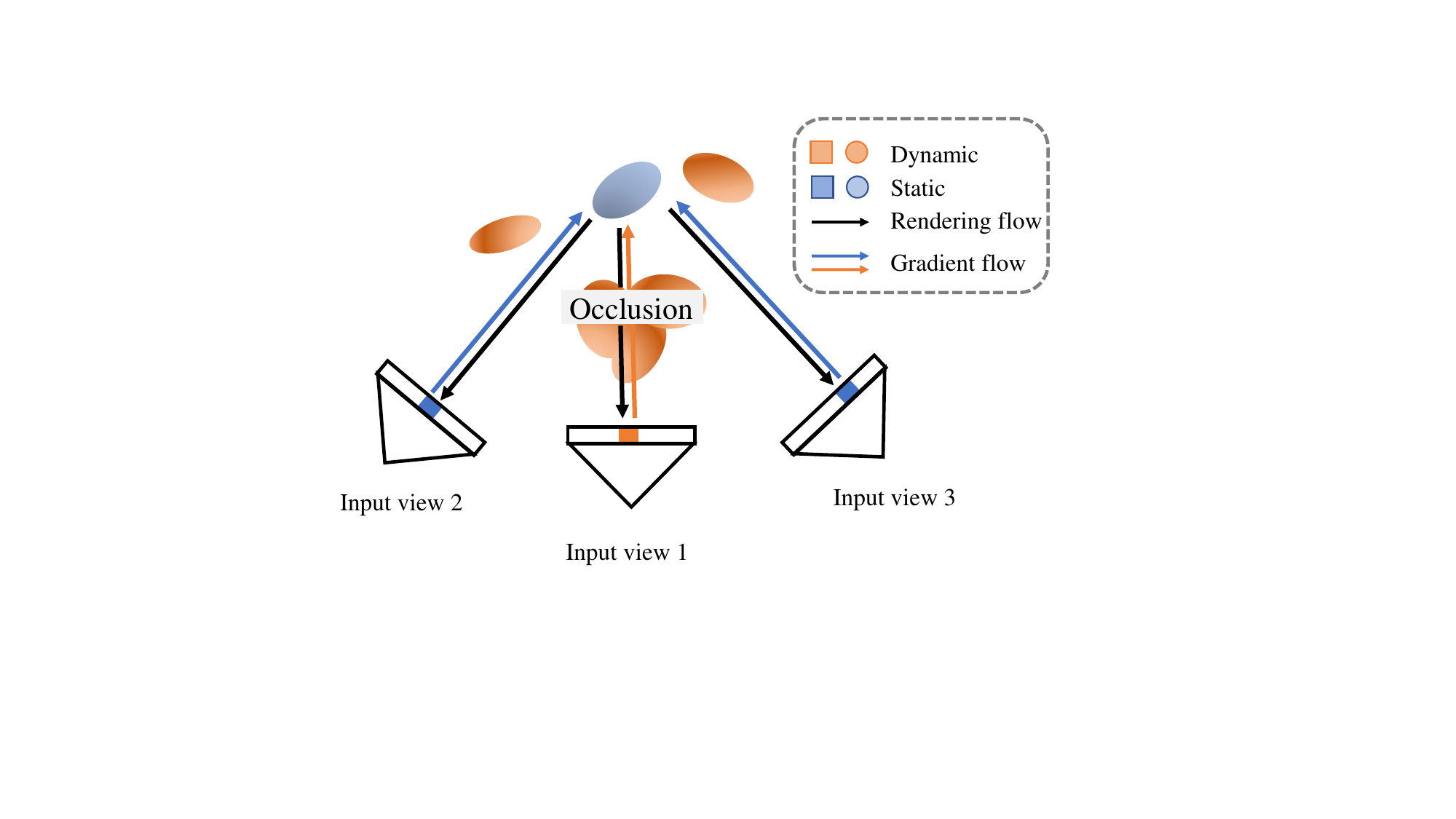}
        \subcaption[]{}
    \end{minipage}
    \hfill
    \begin{minipage}{0.5\textwidth}
        \centering
        \begin{minipage}{0.49\textwidth}
            \centering
            \includegraphics[width=\textwidth]{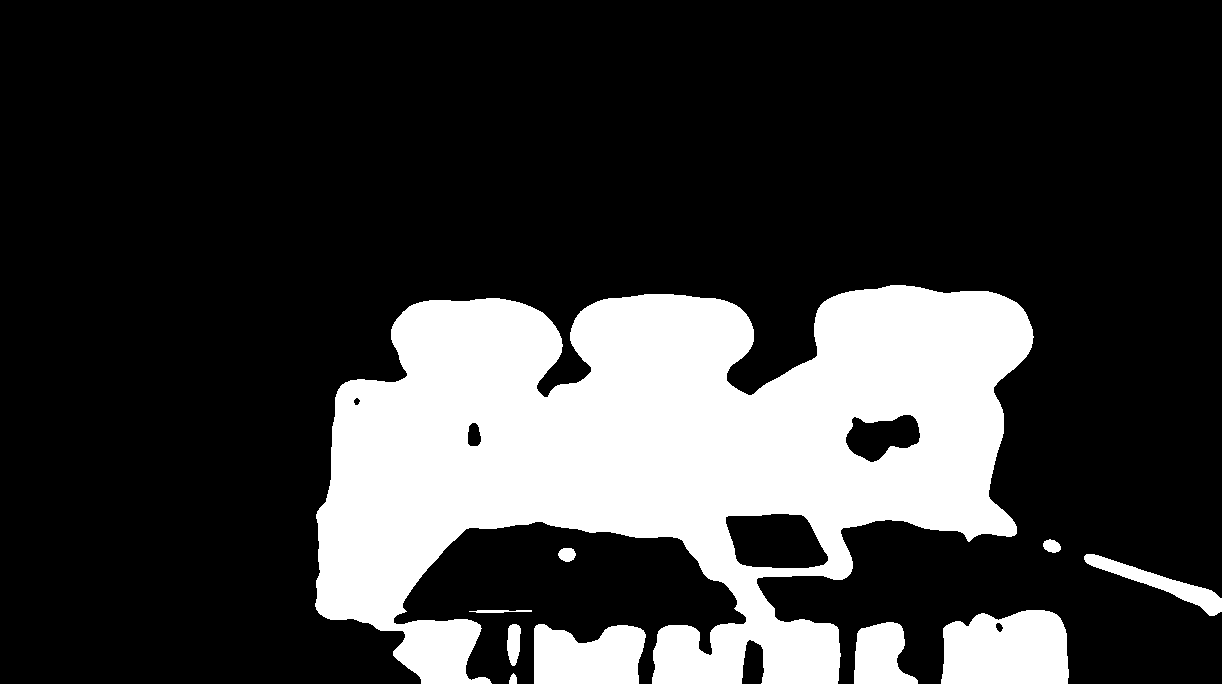}
        \end{minipage}
        \hfill
        \begin{minipage}{0.49\textwidth}
            \centering
            \includegraphics[width=\textwidth]{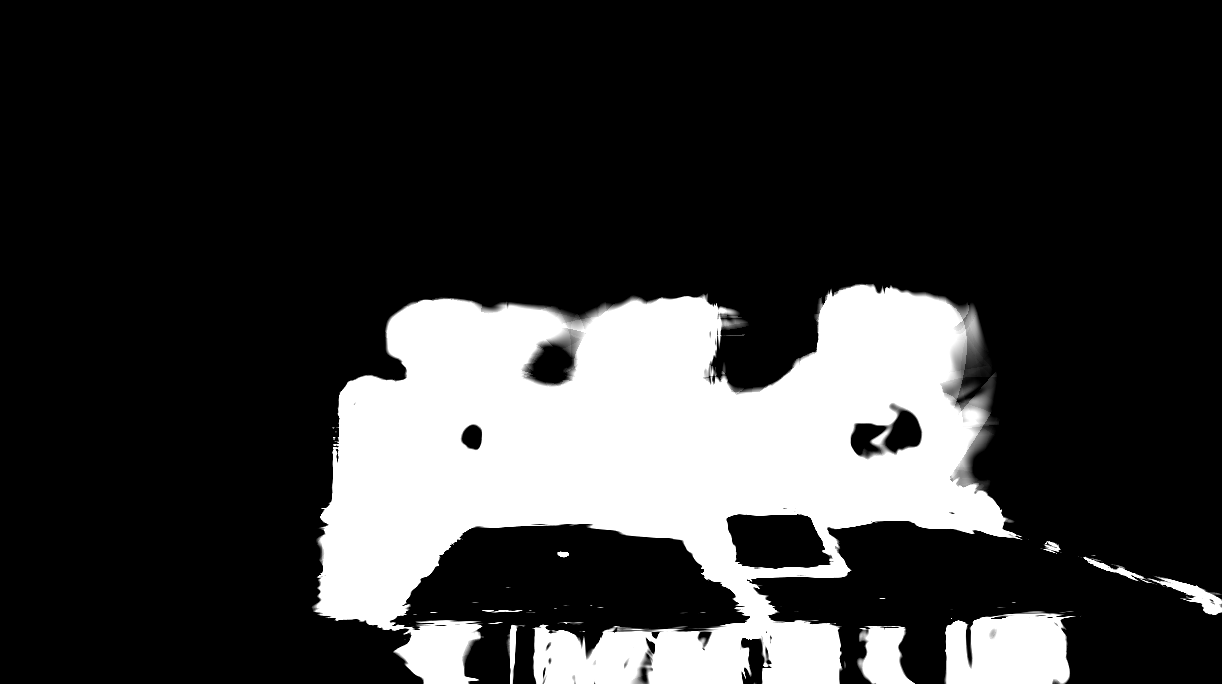}
        \end{minipage}
        \vfill
        \begin{minipage}{0.49\textwidth}
            \centering
            \includegraphics[width=\textwidth]{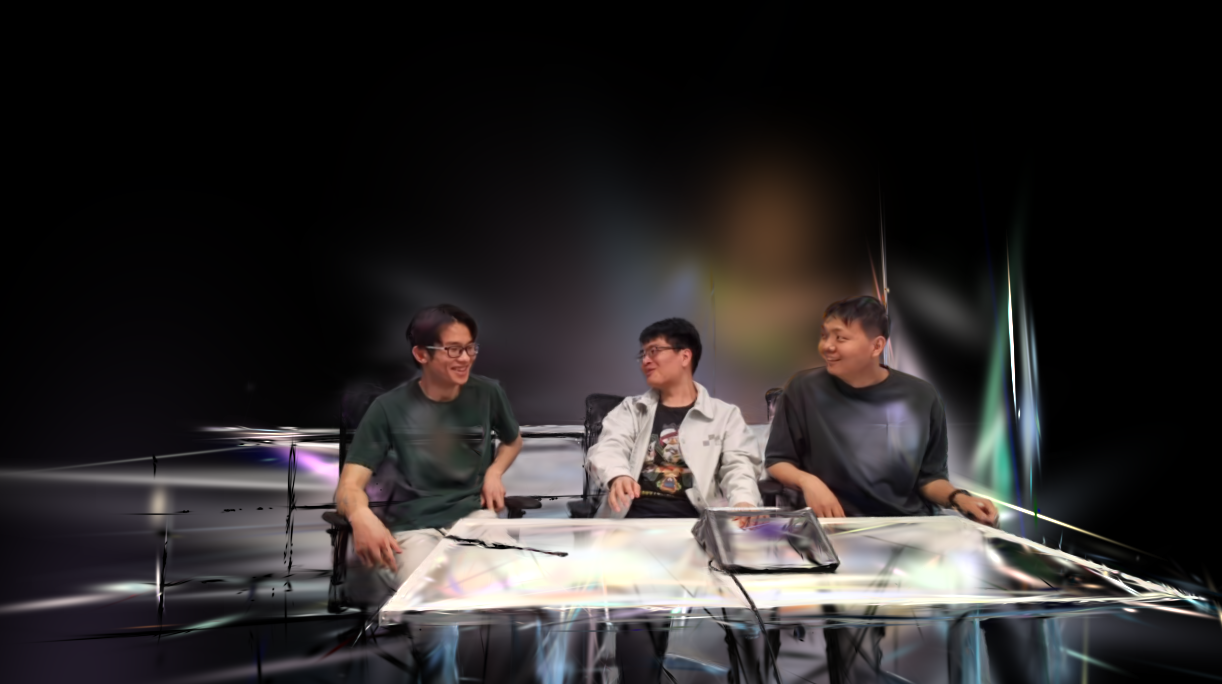}
        \end{minipage}
        \hfill
        \begin{minipage}{0.49\textwidth}
            \centering
            \includegraphics[width=\textwidth]{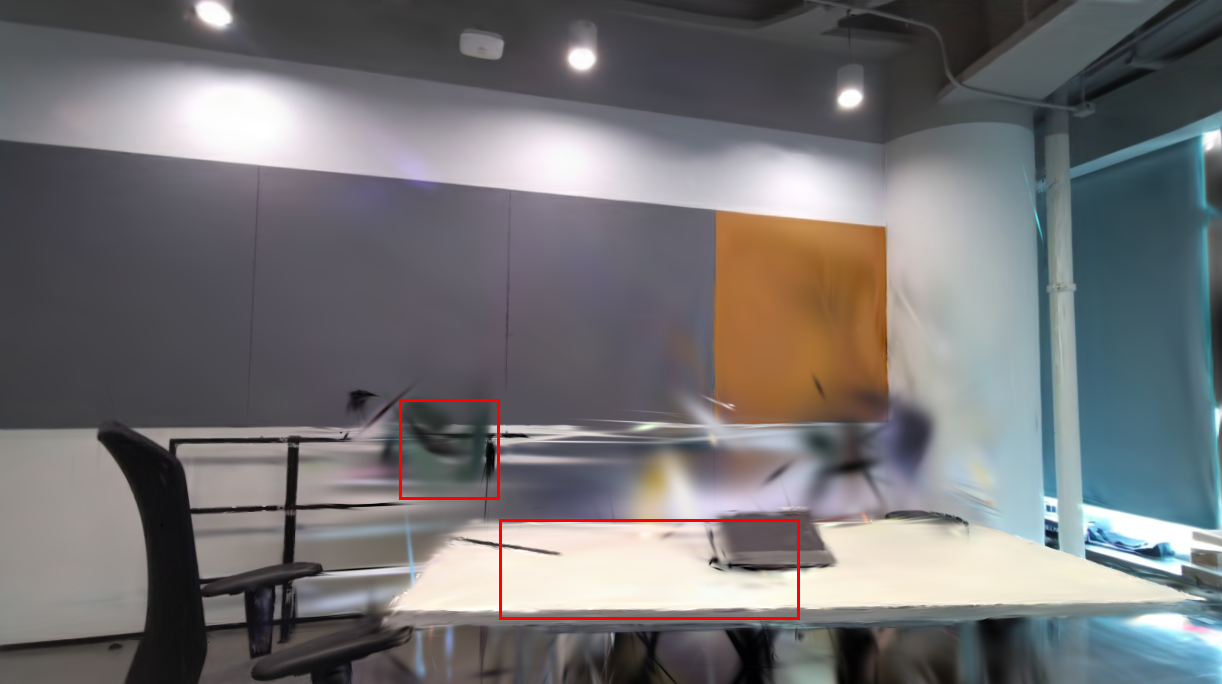}
        \end{minipage}
        \subcaption[]{}
    \end{minipage}
    
    \caption{(a) Diagram illustrating dynamic parameter \(d\) optimization. Even when static points (blue) are occluded by dynamic points (orange) from View 1, they can still be correctly optimized from View 2 and 3. (b) shows the result of decomposition. From top left to bottom right, the order is GT mask, dynamic parameter rendered image, dynamic point and static point rendering results.}
    \label{figs:dyn and sta}
\end{figure}

\subsection{Efficient dynamic and static decomposition}
\label{sec:segmentation}


In this section, we introduce our dynamic-static decomposition method for eliminating redundant computations for static Gaussian points. The time-varying motion model is applied solely to the dynamic components, leaving the static elements unchanged. This approach leads to faster convergence and enhanced rendering quality. Specifically, we introduce a learnable dynamic parameter \( d \) (Initialized to 0.) within the Gaussian points to quantify the the dynamic level of each points. A higher \( d \) corresponds to more pronounced motion, indicating that the point is likely dynamic. We first compute a 2D dynamic-static pixel mask \( D(x) \) from the training videos to distinguish dynamic and static pixels, as shown in Eq. \ref{eq:3}, which serves as the supervision signal.
\begin{equation}
\label{eq:3}
  \left.D(x)=\left\{\begin{array}{ll}1&\quad  S(x) >= \gamma  \\0&\quad  S(x) < \gamma \end{array}\right.\right.  
  ~~~ \text{where} ~~~     
  S(x)=\sqrt{\frac{1}{T}\sum_{t=1}^T(C(x,t)-\frac{1}{T}\sum_{t=1}^TC(x,t))^2}
\end{equation}

where C(x, t) represents the pixel intensity of in \(t\) th frame at location  $x \in \mathds{R}^2$. \( S(x) \) is the temporal standard deviation (std) for each pixel \( x \) across the entire time duration \( T \). 
Subsequently, a threshold of $\gamma = 0.02$ is applied to binarize \(S(x)\), generating a pixel mask \(D(x)\). Pixels with \(S(x)\) greater than or equal to $\gamma$ are classified as dynamic, while those below are considered static.

Based on the concept: During backpropagation, Gaussian points intersecting dynamic pixels should receive a positive gradient, while those intersecting static pixels should receive a negative gradient, with the gradient gradually weakening with distance and occlusion.
Our decomposition design is illustrated in Fig. \ref{figs:dyn and sta}(a). We use the \(\alpha\) composition for parameter \( d \) with Sigmoid function to render a dynamic value \( \hat D(x) \) at location x, as Eq. \ref{eq:4} shows.
\begin{equation}
\label{eq:4}
    \hat D(x)=\text{Sigmoid}( \sum_{i=1}d_i \alpha_i(x) \prod_{j=1}^{i-1}(1-\alpha_j(x)) )
\end{equation}
By applying the \textbf{sigmoid} function, we optimize the dynamic parameter \(d\) to span \( (-\infty, +\infty) \), enabling finer differentiation of dynamic degrees. This approach converts our dynamic-static decomposition into a binary classification problem. Consequently, optimizing the dynamic value for each Gaussian point can be accomplished by minimizing the binary cross-entropy loss:
\begin{equation}
    \mathcal{L}_d=  \mathbb{E}_x [ - D(x)\mathrm{log}(\hat{D}(x))-(1-D(x))\mathrm{log}(1-\hat{D}(x))].
\end{equation}
From the equation above, we effectively optimize the dynamic parameter \(d\) for the Gaussian points. The entire optimization process is highly efficient, typically concluding within 1 minute. Ultimately, Gaussian points with dynamic parameter greater than the dynamic threshold \( \zeta = 7.0 \)  are classified as dynamic; otherwise, they are classified as static. An ablation study on \(\zeta\) is presented in Fig. \ref{fig:curve} and Sec. \ref{Sec:Discussion}, demonstrating that our decomposition method is robust to the choice of \(\zeta\).

Notably, our method adapts well to occlusion. In Fig. \ref{figs:dyn and sta}(a), while dynamic pixels (orange) from View 1 incorrectly assign positive dynamic values to static Gaussian points (blue), other views like View 2 and 3 assign larger negative values, ensuring correct classification.  

\subsection{Spatio-temporal structure}
\label{sec:4dhash}
We introduce our proposed efficient spatio-temporal structure encoder, the 4D multi-resolution hash \( h_{4d} \), and the deformation decoder MLPs, used to predict the deformation of each dynamic Gaussian. 

\textbf{4D Multi-resolution hash encoder.} Inspired by INGP \cite{muller2022instant}, we propose utilizing a 4D multi-resolution hash \( h_{4d} \) for encoding to effectively model the temporal information of dynamic Gaussians by normalizing the point cloud into the hash grid range. As described in INGP, voxel grid at each resolution is mapped to a hash table that stores \(F\)-dimensional learnable feature vectors. For a given 4D dynamic Gaussian \((\mu, t) \in \mathbb{R}^4\), its hash encoding at resolution \(l\), denoted as \(h_{4d}(\mu, t; l) \in \mathbb{R}^F\), is computed through linear interpolation of the feature vectors associated with corners of the surrounding grid. Consequently, its multi-resolution hash encoding features are as follows:
\begin{equation}
    f_h =  [h_{4d}(\mu,t;0),h_{4d}(\mu,t;1)...h_{4d}(\mu,t;L-1)] \in  \mathbb{R}^{LF},
\end{equation}
where \( L \) denotes the number of resolution levels, typically set to 16. Following this, a small MLP \( \phi_d \) combines all features to produce \( f_d = \phi_d(f_h) \). 
Using the 4D hash \( h_{4d} \) as an encoder offers several advantages: compactness, \( O(1) \) query complexity, and the multi-resolution approach effectively integrates global and local information. 

However, while 4D Hash offers \(O(1)\) query complexity, its hashing characteristics make encoding the temporal information of an entire scene both challenging and storage-intensive. Fortunately, our proposed decomposition method focuses on encoding the temporal information of dynamic points only, reducing the need for a larger hashing space and simplifying the modeling of the scene's temporal domain.
This approach enables us to retain the fast access speed of 4D Hash while minimizing storage requirements.


\textbf{Multi-head Gaussian Deformation Decoder.} Once all features of dynamic Gaussian points are encoded, we can compute any required variables using a multi-head Gaussian deformation decoder MLPs = \{ \(\phi_\mu,\phi_s,\phi_q,\phi_\sigma,\phi_{sh} \) \}:
\begin{equation}
     \text{d}\mu, \text{d}s, \text{d}q, \text{d}\sigma, \text{d}sh = \text{MLPs}(f_d) 
\end{equation}
Here, \( \text{d}\mu, \text{d}s, \text{d}q, \text{d}\sigma, \text{d}sh \) represent the deformation intensity of the mean, scaling, rotation, opacity, and color of the Gaussian point at time \( t \). Therefore, the deformed parameters of dynamic Gaussian \( G_d \) can be expressed as:
\begin{equation}
     (\mu', r', q', \sigma', sh' )= (\mu + \text{d}\mu, r + \text{d}r, q + \text{d}q, \sigma + \text{d}\sigma, sh + \text{d}sh )
\end{equation}
where ( \(\mu', r', q', \sigma', sh'\) ) represent the new parameters of the dynamic Gaussian at time \( t \). For static Gaussian elements \( G_s \), they are directly combined with the deformed dynamic Gaussian elements \( G_d \) to render the final rendered image \( I_t \).


\subsection{Density control}
\label{sec:densification and pruning}
In the original 3DGS, the opacity of all points is regularly reduced, and Gaussian points with low transparency are clipped during the pruning stage. However, this approach is not appropriate for our method as it results in excessive coupling between the canonical space and the deformation space. Therefore, we eliminate the reset opacity operation. 
Inspired by previous works \cite{niemeyer2024radsplat,deng2024compact,fan2023lightgaussian}, which focus on the compact representation of static scenes by pruning redundant Gaussians based on spatial attributes such as transparency and volume. We adopt a novel approach to pruning floaters across canonical and deformation space: Temporal Importance Pruning, as shown in Fig. \ref{figs:pipelins}. This involves calculating the importance of each Gaussian point to each training viewpoint at every timestamp. Gaussians with importance below a certain threshold can be clipped, effectively reducing floater issues. For a Gaussian point \( g_i \), the importance \(w_i\) is calculated as follows:                     
\begin{equation}
    w_i=\max_{I\in\mathcal{I},x\in I,t\in T} (\alpha_i(x | t) \prod_{j=1}^{i-1}(1-\alpha_j(x | t))) 
\end{equation}
Here, \( \mathcal{I} \) represents the images from all training views, \(\alpha_i( x | t)\) is the value of \(\alpha_i(x)\) at time \(t\) in Eq. \ref{eq:blending}, \( T \) represents the set of query times. We prune Gaussians when their importance satisfies \( w_i < 0.02 \). 
As illustrated in Fig. \ref{fig: ablation prune}, this method effectively eliminates artifacts that are suspended in the air and were not captured by the training views.  For the cloning and splitting of Gaussians, we adhere to the procedures of 3DGS, with the child Gaussians inheriting the dynamic properties of their parent Gaussians.

\begin{table}[t]
    \centering
    \vspace{-2em}
    \caption{\textbf{Quality comparison on the N3DV dataset.} The \textcolor{red!50}{best} and the \textcolor{blue!50}{second best} results are denoted by red and blue. $^{1}$ online method.}
    \label{tab: nv3d all quality comparison}
    \begin{threeparttable}
    \resizebox{\textwidth}{!}
    {
        \begin{tabular}{ccccccccccccc}
        \toprule[1.5pt]
         Method & PSNR \(\uparrow\)  & DSSIM \(\downarrow\)  & LPIPS \(\downarrow\)     & Time \(\downarrow\)  & Size(MB) \(\downarrow\) & FPS \(\uparrow\) \\

        \midrule[0.5pt]
        \multirow{1}{*}{DyNeRF \cite{li2022neural}} & 29.58 & 0.020 & 0.099 & 1300.0 hours & \cellcolor{red!25}{30} & 0.02 \\
        
        \multirow{1}{*}{NeRFPlayer \cite{song2023nerfplayer}} & 30.69 & - & 0.111 & 6.0 hours & 5100 & 0.05 \\

        \multirow{1}{*}{HexPlane \cite{cao2023hexplane}} & 31.70 & \cellcolor{red!25}{0.014} & 0.075 & 12.0 hours & 240 & 0.21 \\

        \multirow{1}{*}{K-Planes \cite{fridovich2023k}} & 31.63 & 0.018 & - & 5.0 hours & 300 & 0.15 \\

        \multirow{1}{*}{4DGS \cite{wu20244d}}  & 31.02 &  0.030 & 0.150 & 50 mins & \cellcolor{blue!25}{90} & 30  \\
        
        \multirow{1}{*}{3DGStream $^1$  \cite{sun20243dgstream}}  & 31.67 &  - & - & 60 mins & 2340 & \cellcolor{red!25}215  \\

        \multirow{1}{*}{SpaceTimeGS \cite{li2024spacetime}}  & \cellcolor{blue!25}{32.05} &  \cellcolor{red!25}{0.014} & \cellcolor{blue!25}{0.044} & 10.0 hours & 200 & 110 \\
        
        \multirow{1}{*}{Real-Time4DGS \cite{yang2023real}}  & 32.01 &  \cellcolor{red!25}{0.014} & 0.055 & 9.0 hours &  \(>1000\) & 114 \\
        \multirow{1}{*}{\textbf{\confDataName Lite(Ours)}}  & 31.79 &  \cellcolor{blue!25}{0.017}  & 0.072  & \cellcolor{red!25}{20 mins}   & \cellcolor{red!25}{30}  &128 \\
        
        \multirow{1}{*}{\textbf{\confDataName(Ours)}}  & \cellcolor{red!25}{32.23} &  \cellcolor{red!25}{0.014} & \cellcolor{red!25}{0.043} & \cellcolor{blue!25}{25 mins}   & 120  & \cellcolor{blue!25} 125 \\
        
        \bottomrule[1.5pt]
        \end{tabular}
    }

\end{threeparttable}
\end{table}

\subsection{Optimization pipeline}
\label{sec:optimization}

We start by initializing the SfM \cite{schonberger2016structure} point cloud using the first frames, then train on the first-frame images for 5000 iterations to establish a well-defined canonical space. Next, training the dynamic attributes of each Gaussian point within the canonical space takes about 1 minute, followed by training the spatio-temporal structure. Consistent with the principles of 3DGS, our loss function remains simple, without additional terms:
\[ \mathcal{L}_{rec} = (1 - \lambda_1)\mathcal{L}_1 + \lambda_1\mathcal{L}_{SSIM} \]

\begin{table}[t]
    \centering
    \caption{\textbf{Quantitative comparison on the MeetRoom dataset.} PSNR is averaged across all frames, while training time and storage requirements accumulate over the entire sequence. $^{1}$ online method.} 
    \label{tab: meetroom all quality comparison}
    \begin{threeparttable}
    \resizebox{0.7\textwidth}{!}
    {
        \begin{tabular}{ccccccccccccc}
        \toprule[1.5pt]
         Method & PSNR \(\uparrow\)  & Time(hours) \(\downarrow\)  & Size(MB) \(\downarrow\) \\
        \midrule[0.5pt]
        \multirow{1}{*}{Plenoxel\cite{fridovich2022plenoxels}}  & 27.15 & 70 & 304500 \\
        \multirow{1}{*}{I-NGP\cite{muller2022instant}}  & 28.10 & 5.5 & 14460 \\
        \multirow{1}{*}{3DGS\cite{kerbl20233d}}  & 31.31 & 13 & 6330 \\
        \midrule[0.5pt]
        \multirow{1}{*}{ StreamRF $^{1}$\cite{li2022streaming}}  & 26.72 & 0.85 & 2700 \\
        \multirow{1}{*}{ 3DGStream $^{1}$\cite{sun20243dgstream}}  & 30.79& 0.6 & 1230 \\
        \multirow{1}{*}{\textbf{\confDataName(Ours)}}  & \cellcolor{red!25}{32.05} & \cellcolor{red!25}{0.3} & \cellcolor{red!25}{40}  \\
        \bottomrule[1.5pt]
        \end{tabular}
    }

\end{threeparttable}
\end{table}

\begin{figure}[t]
\centering
\vspace{-2em}
\setcounter{subfigure}{0}  
\subfloat[GT]{
		\includegraphics[scale=0.078]{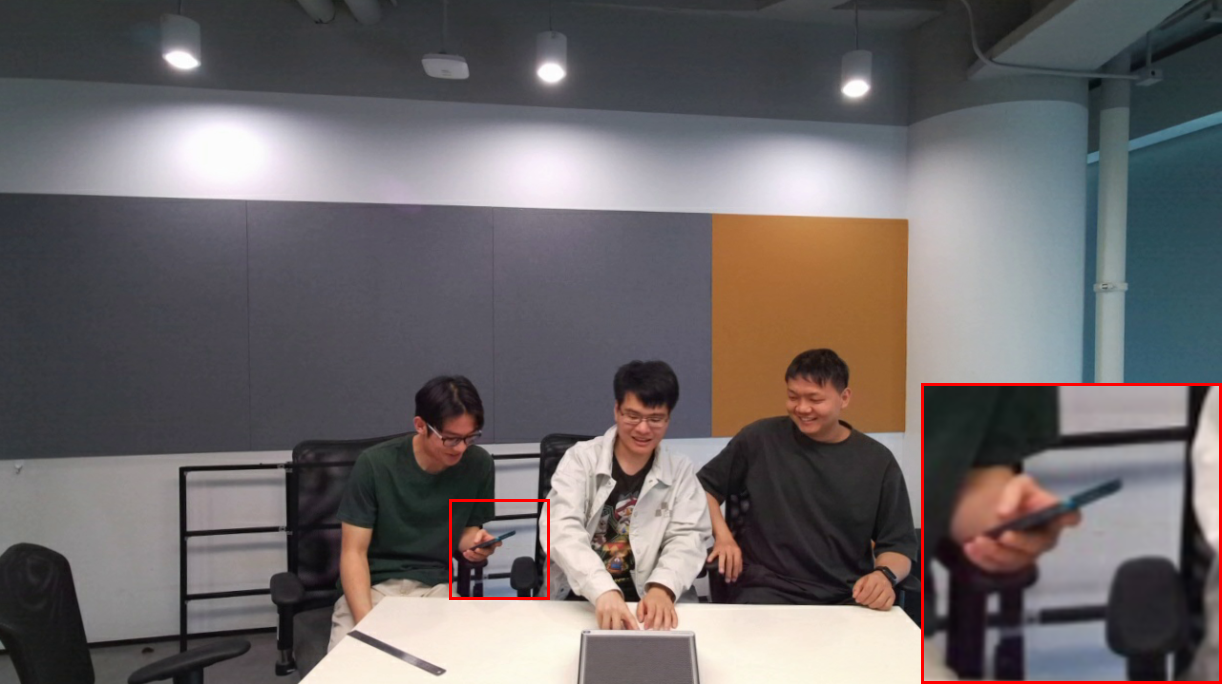}}
  \subfloat[Ours]{
		\includegraphics[scale=0.078]{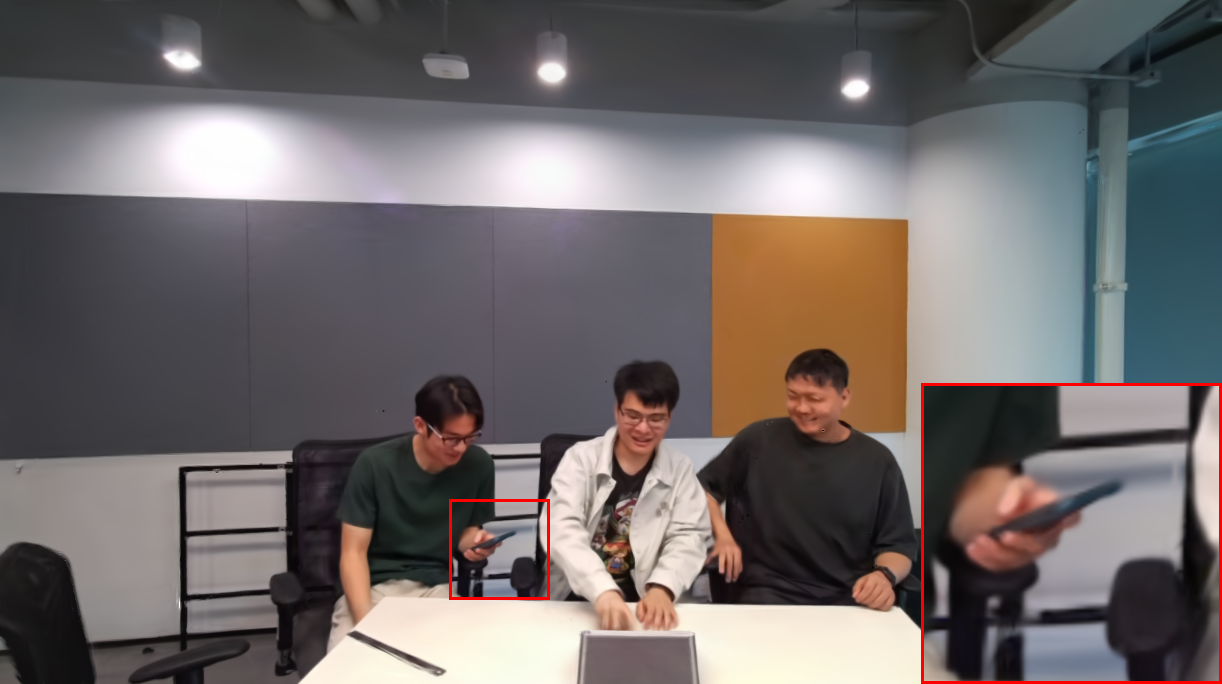}}
\subfloat[3DGStream]{
		\includegraphics[scale=0.078]{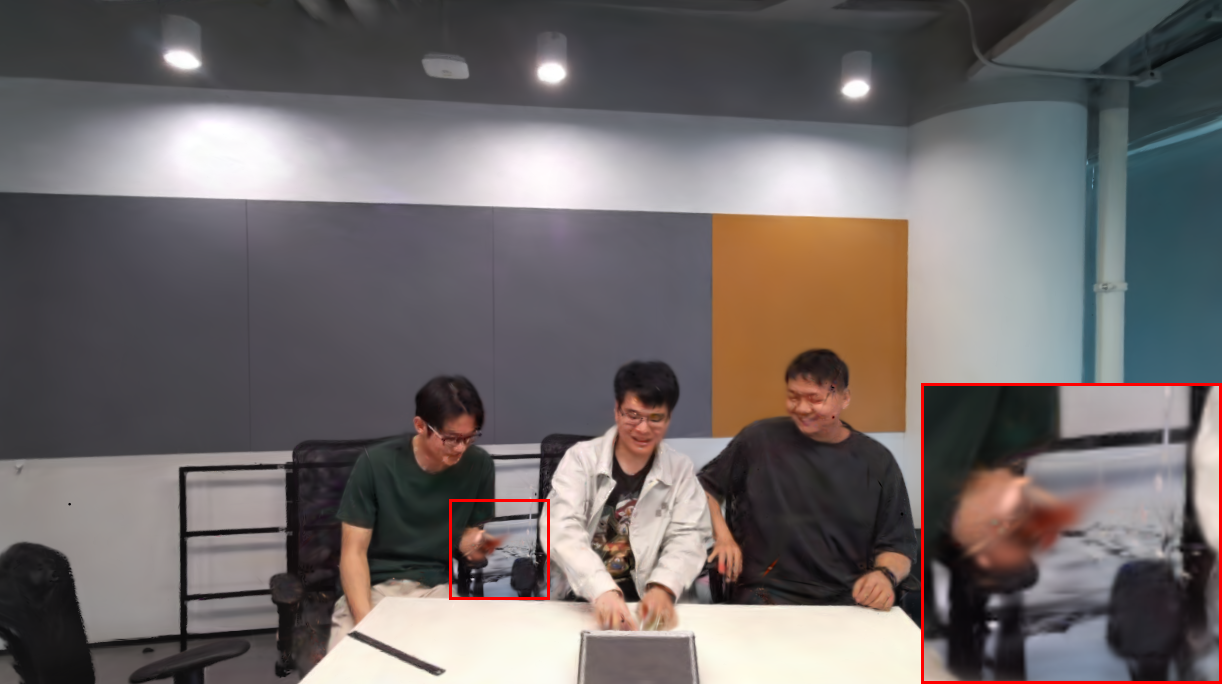}}
\subfloat[3DGS]{
		\includegraphics[scale=0.078]{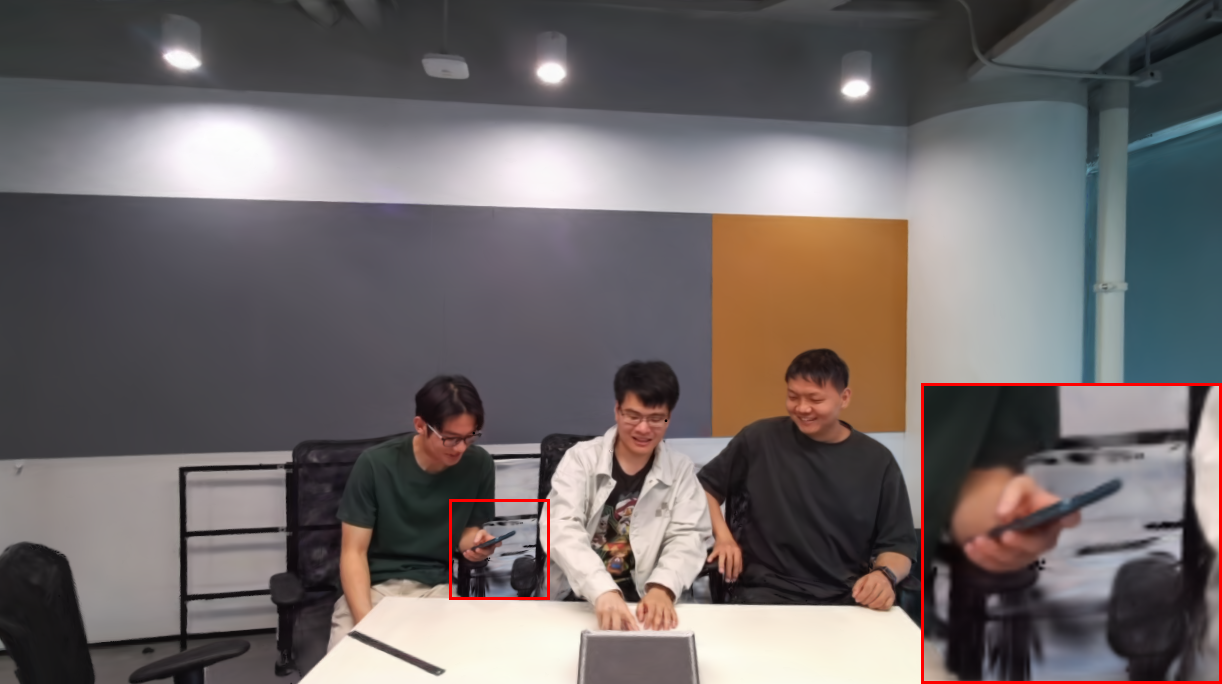}}
\caption{ Qualitative result on the \textit{discussion}.}
\label{fig: meetroom compairl}
\vspace{-1em}
\end{figure}

\section{Experiment}
\label{sec:experment}

In this section, we provide details of our implementation and datasets in Sec. \ref{sec:details} and \ref{sec:dataset}, respectively. A thorough analysis of the experimental results is presented in Sec. \ref{sec:evaluation}, while Sec. \ref{sec:ablation} covers the ablation experiments for our method. The results show that our approach achieves sota performance in terms of training speed, storage efficiency, and rendering quality.

\subsection{Implementation details}
\label{sec:details}
We initialize with point clouds generated by Colmap, followed by constructing our canonical space using the first frames from all training viewpoints, trained for 5000 epochs. Next, we train the dynamic parameter \( d \) of Gaussian points using the Adam optimizer \cite{kingma2014adam} with a learning rate of 0.05. This training spans 3000 epochs and completes in under 1 minute. Finally, we train our spatio-temporal structure for approximately 14000 epochs, utilizing settings for the 4D Hash table similar to those in InstantNGP \cite{muller2022instant}. We use the Adam optimizer with an initial learning rate of 0.002, which exponentially decays to 0.00002 over the course of training. Lite refers to a lite-version model with a hash table size set to \(2^{15}\) and \(\lambda_1 = 0\).  All experiments were conducted on an NVIDIA RTX 3090 GPU.

\subsection{Dataset}
\label{sec:dataset}

\begin{wraptable}{r}{7cm}
\vspace{-1em}
    \centering
    \caption{ Quantitative results comparison for \textit{sear steak} and \textit{flame steak} includes average PSNR and time metrics over 300 frames of the test view. }
    \label{tab: ablation exp}
    \begin{threeparttable}
\resizebox{0.5\textwidth}{!}
{
\begin{tabular}{ccccccccccccc}
    \toprule[1.5pt]
     Method & PSNR \(\uparrow\)  & Time(mins) \(\downarrow\)  \\
    \midrule[0.5pt]
    \multirow{1}{*}{Lite}  & 33.31 & 20 \\
    \multirow{1}{*}{Muti-planes}  & 33.48 & 28 \\
    \multirow{1}{*}{W/o decomp.}  & 32.68 & 35 \\
    \multirow{1}{*}{Full}  & 33.83 & 25 \\
    \bottomrule[1.5pt]
    \end{tabular}
}
\end{threeparttable}
\end{wraptable}

\textbf{The N3DV dataset \cite{li2022neural}} is captured using a multi-view system with 18-21 cameras, recording dynamic scenes at a resolution of \(2704 \times 2028\) and 30 FPS. It includes various complex scenarios such as fire, reflections, and new objects. Following prior works \cite{li2022neural,cao2023hexplane,wu20244d,yang2023real,li2024spacetime}, we downsampled the videos by a factor of two and used the same training and testing data splits as established by them.

\textbf{The Meet Room dataset \cite{li2022streaming}} is captured using a multi-view system with 13 cameras, recording dynamic scenes at a resolution of \(1280 \times 720\) and 30 FPS. Following prior works \cite{sun20243dgstream,li2022streaming}, we used 12 views for training and reserved 1 view for testing.

\textbf{The Basketball court dataset \cite{VRU}} is captured using a multi-view system with 34 cameras, recording dynamic scenes at a resolution of \(1920 \times 1080\) and 25 FPS. This dataset encompasses a large scene with many complex situations, including bouncing, fast motion, occlusion, and transient objects, making it highly challenging.

\begin{figure}[t]
    \centering
    \newcommand{\comparisonimage}[1]{%
        \subfloat{\includegraphics[width=0.99\linewidth]{#1_rect3.png}} \\
        \subfloat{\includegraphics[width=0.33\linewidth]{#1_zoom1.png}}
        \subfloat{\includegraphics[width=0.33\linewidth]{#1_zoom2.png}}
        \subfloat{\includegraphics[width=0.33\linewidth]{#1_zoom3.png}}
    }
    \begin{minipage}[b]{0.25\textwidth}
        \centering
        \comparisonimage{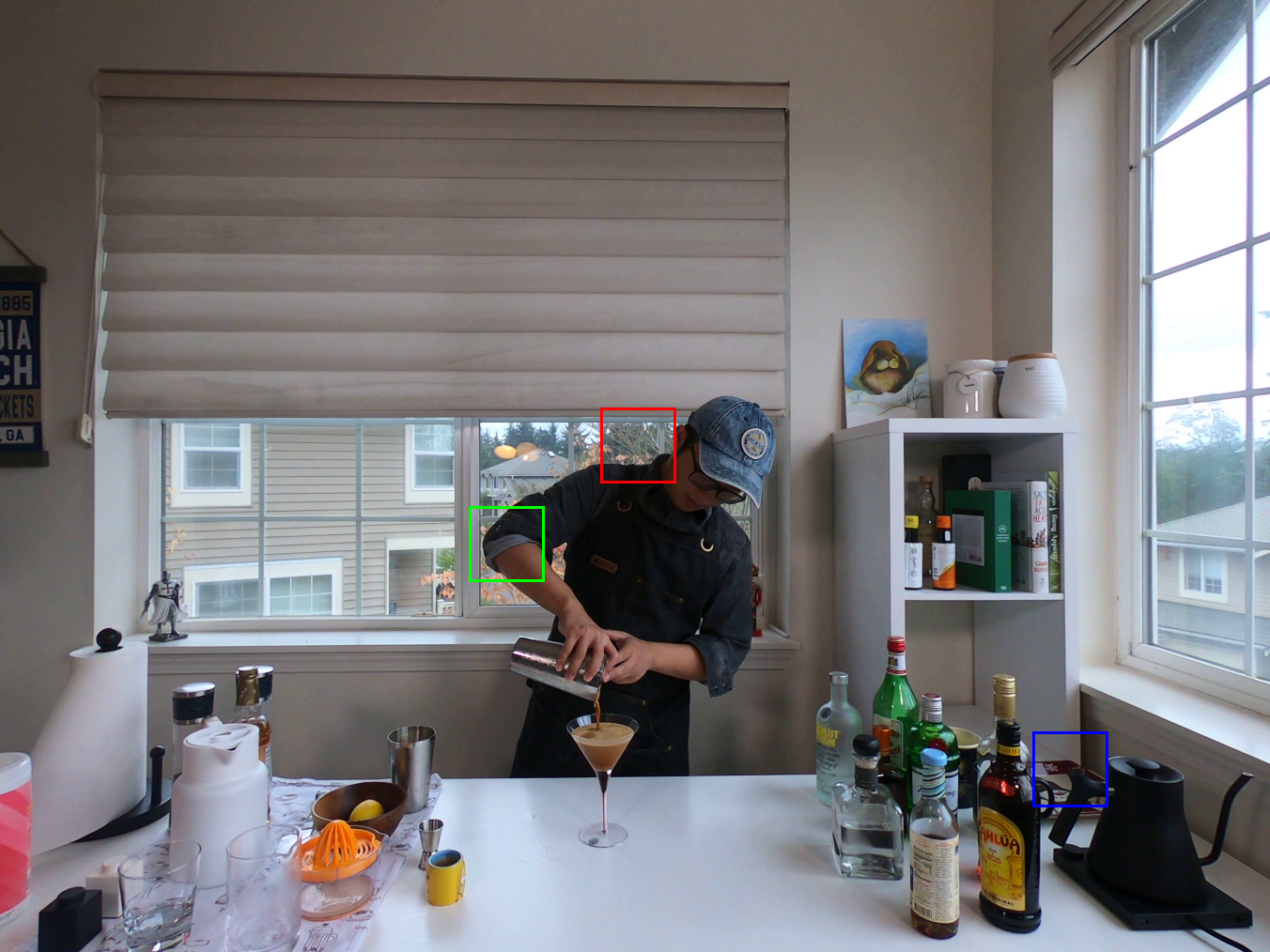}
    \end{minipage}%
    \hfill
    \begin{minipage}[b]{0.25\textwidth}
        \centering
        \comparisonimage{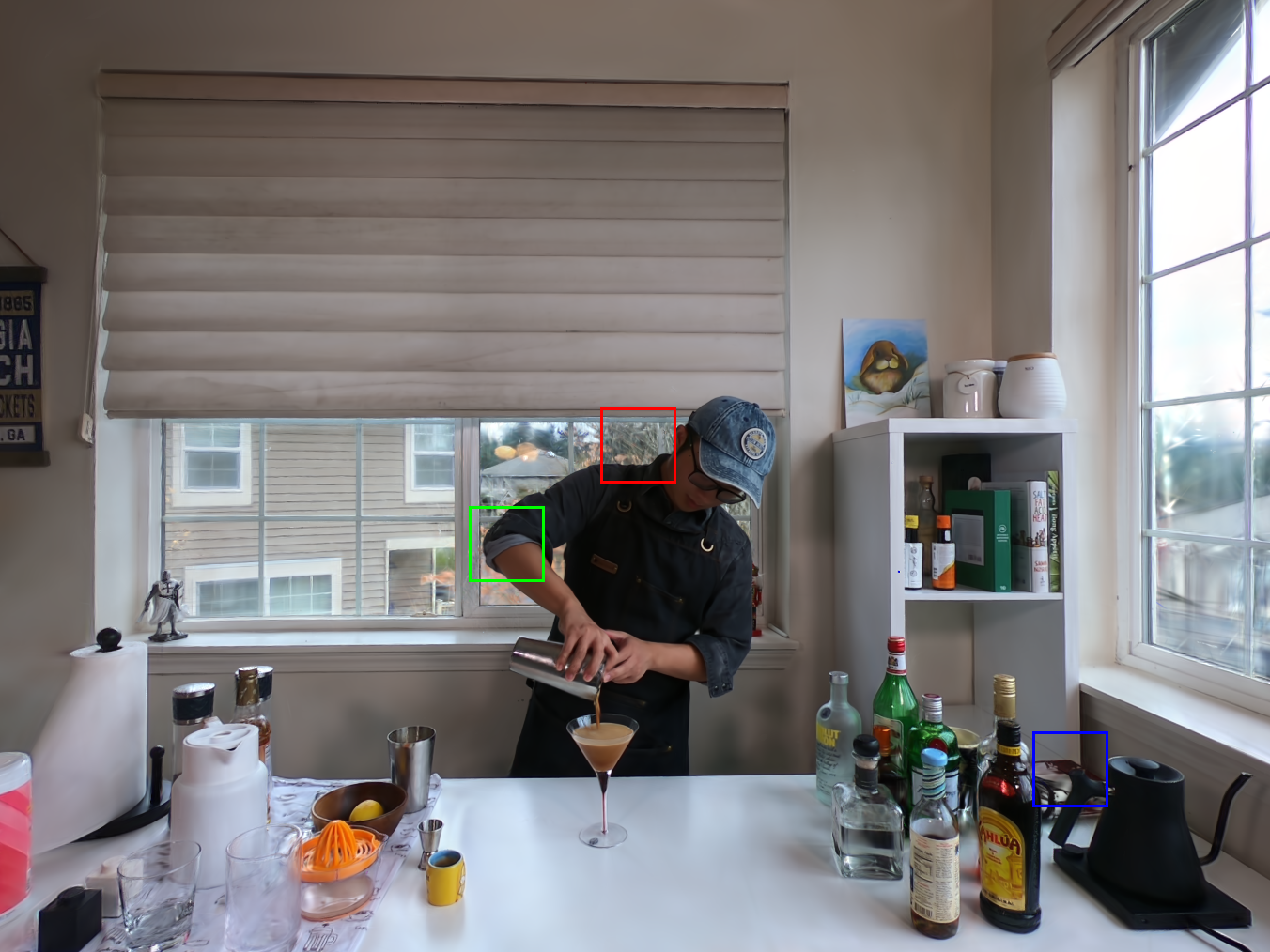}
    \end{minipage}%
    \hfill
    \begin{minipage}[b]{0.25\textwidth}
        \centering
        \comparisonimage{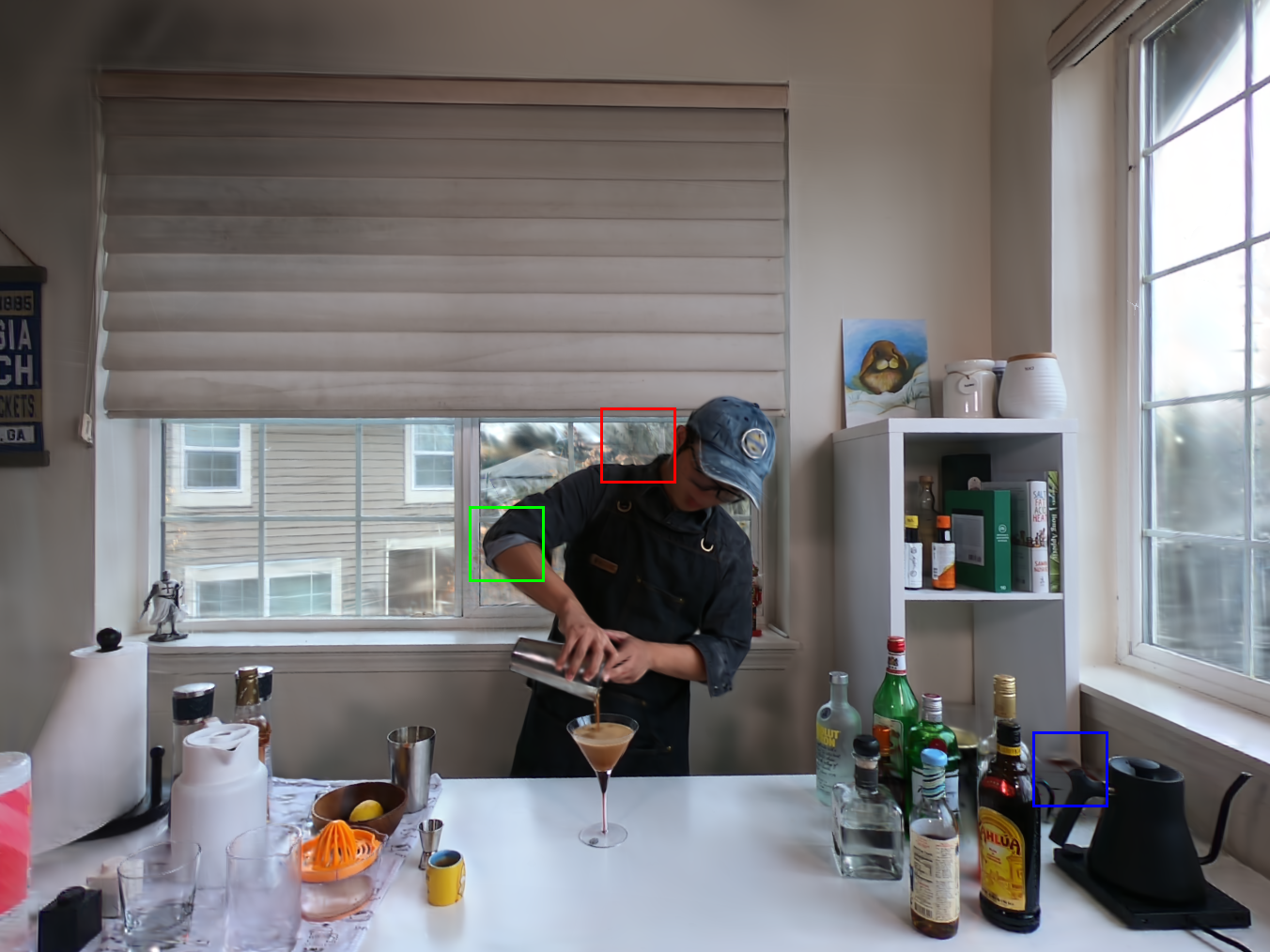}
    \end{minipage}%
    \hfill
    \begin{minipage}[b]{0.25\textwidth}
        \centering
        \comparisonimage{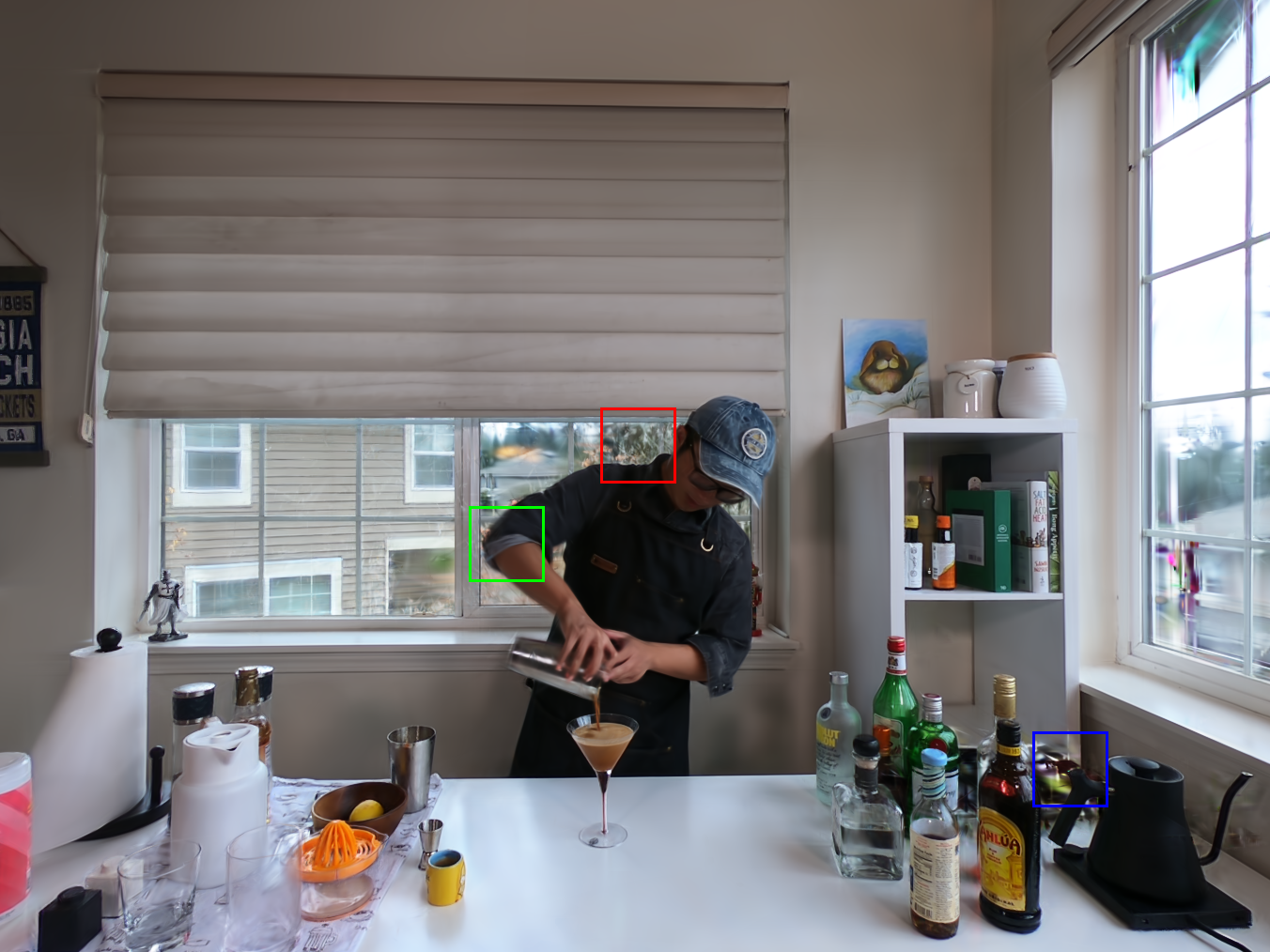}
    \end{minipage}

    \begin{minipage}[b]{0.25\textwidth}
        \centering
        \comparisonimage{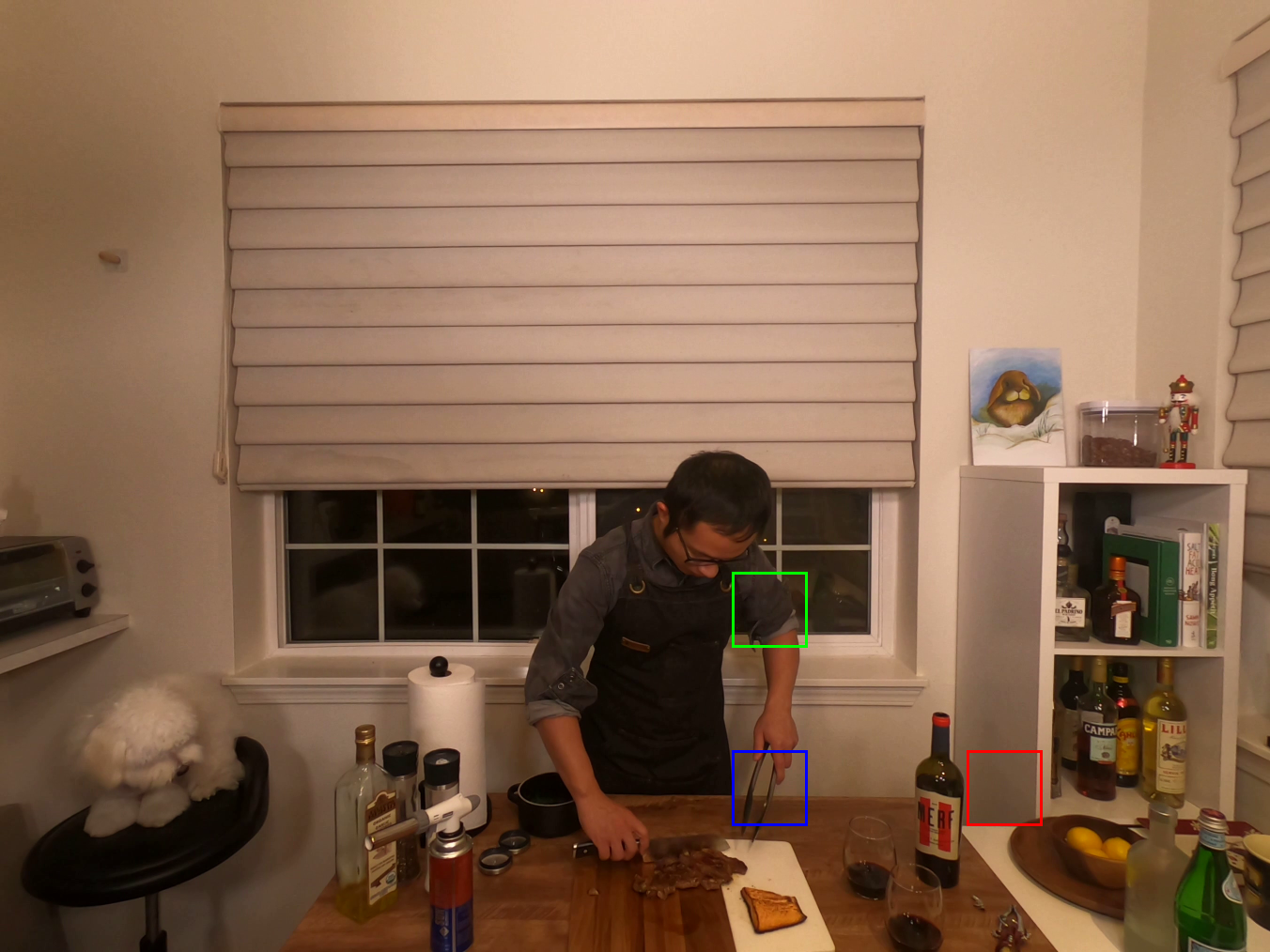}
        \caption*{(a) GT}
    \end{minipage}%
    \hfill
    \begin{minipage}[b]{0.25\textwidth}
        \centering
        \comparisonimage{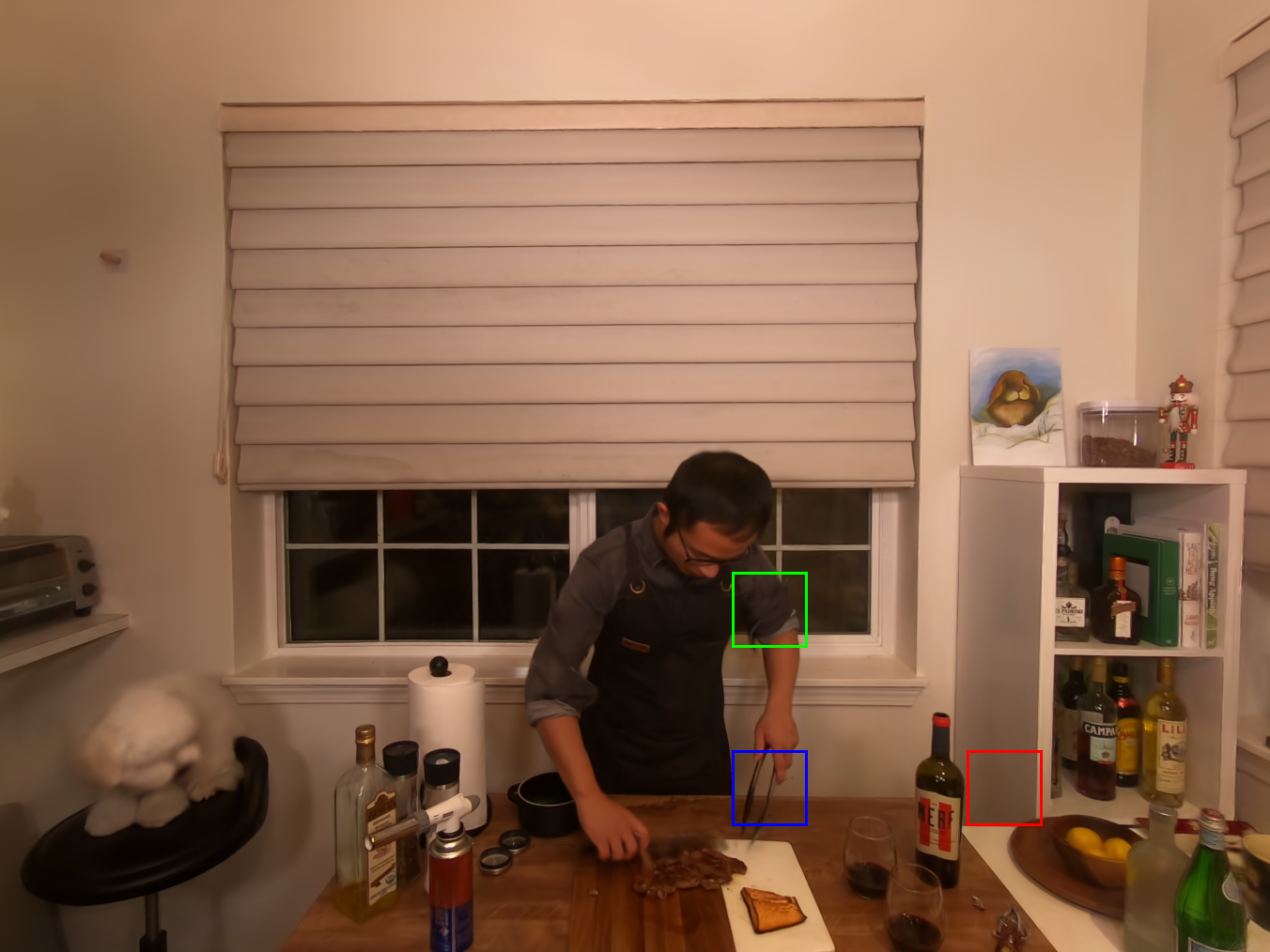}
        \caption*{(b) Ours}
    \end{minipage}%
    \hfill
    \begin{minipage}[b]{0.25\textwidth}
        \centering
        \comparisonimage{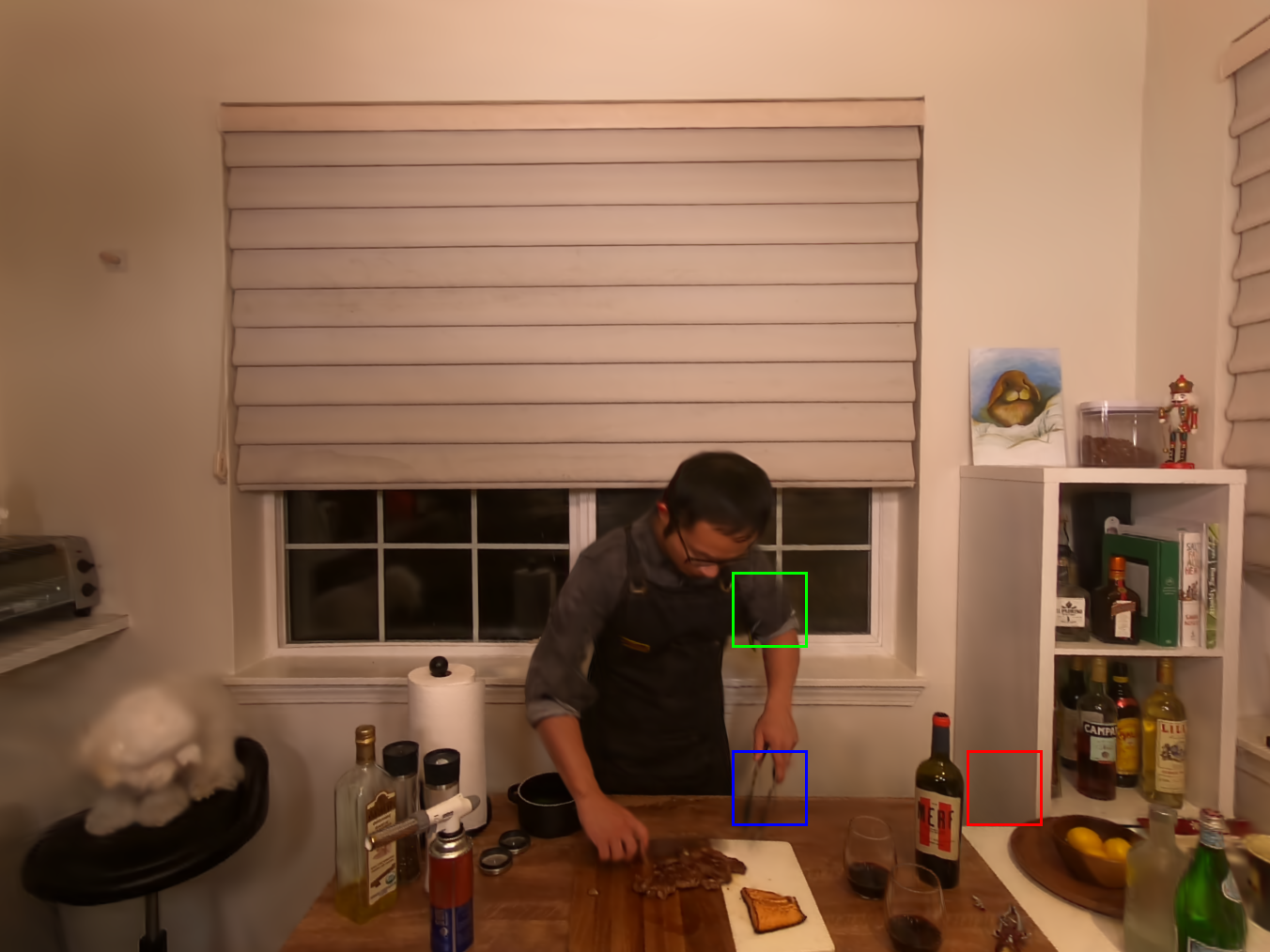}
        \caption*{(c) RTGS}
    \end{minipage}%
    \hfill
    \begin{minipage}[b]{0.25\textwidth}
        \centering
        \comparisonimage{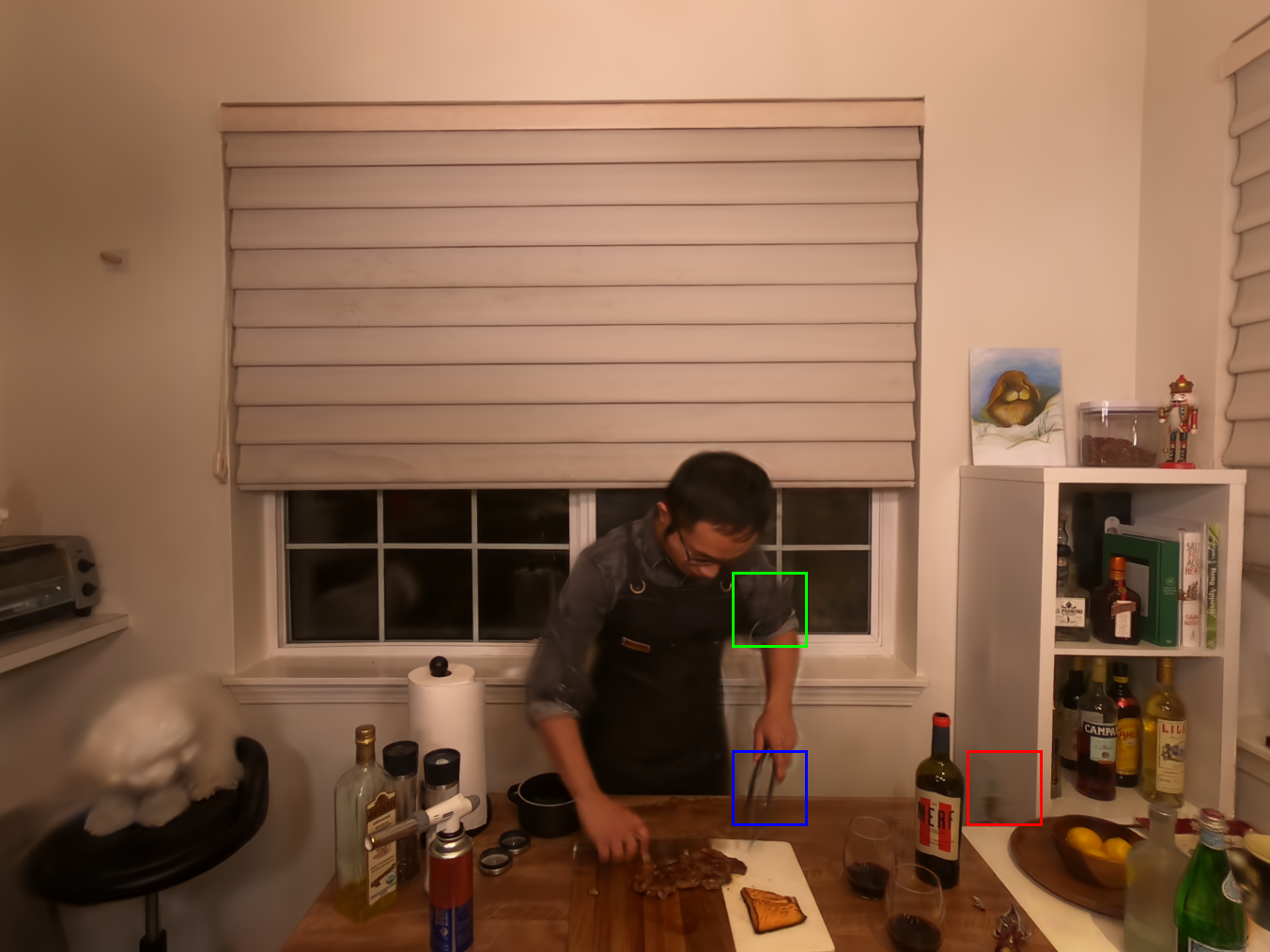}
        \caption*{(d) STGS}
    \end{minipage}
    
    \caption{Qualitative result on \textit{coffee martini} and \textit{cut beef}. It can be observed that our method achieves higher-quality modeling in both dynamic and static regions. }
    \label{fig: nv3d comparison}
    \vspace{-1em}
\end{figure}

\subsection{Evaluation}
\label{sec:evaluation}
For the Meetroom and Basketball court dataset, we follow the processing approach from \cite{sun20243dgstream}, using COLMAP \cite{schonberger2016structure} to estimate the camera pose of the first frame as the global pose. For the N3DV dataset, we adopt the processing approach from \cite{wu20244d}. We evaluate the methods using three metrics across all 300 frames: 1) Average PSNR, DSSIM, and LPIPS \cite{zhang2018unreasonable} scores for the test views; 2) Total training time and FPS; 3) Model size .

\begin{figure}[t]
    \centering
    \vspace{-1em}
    \includegraphics[width= \textwidth ,trim=0cm 14cm 0cm 0cm]{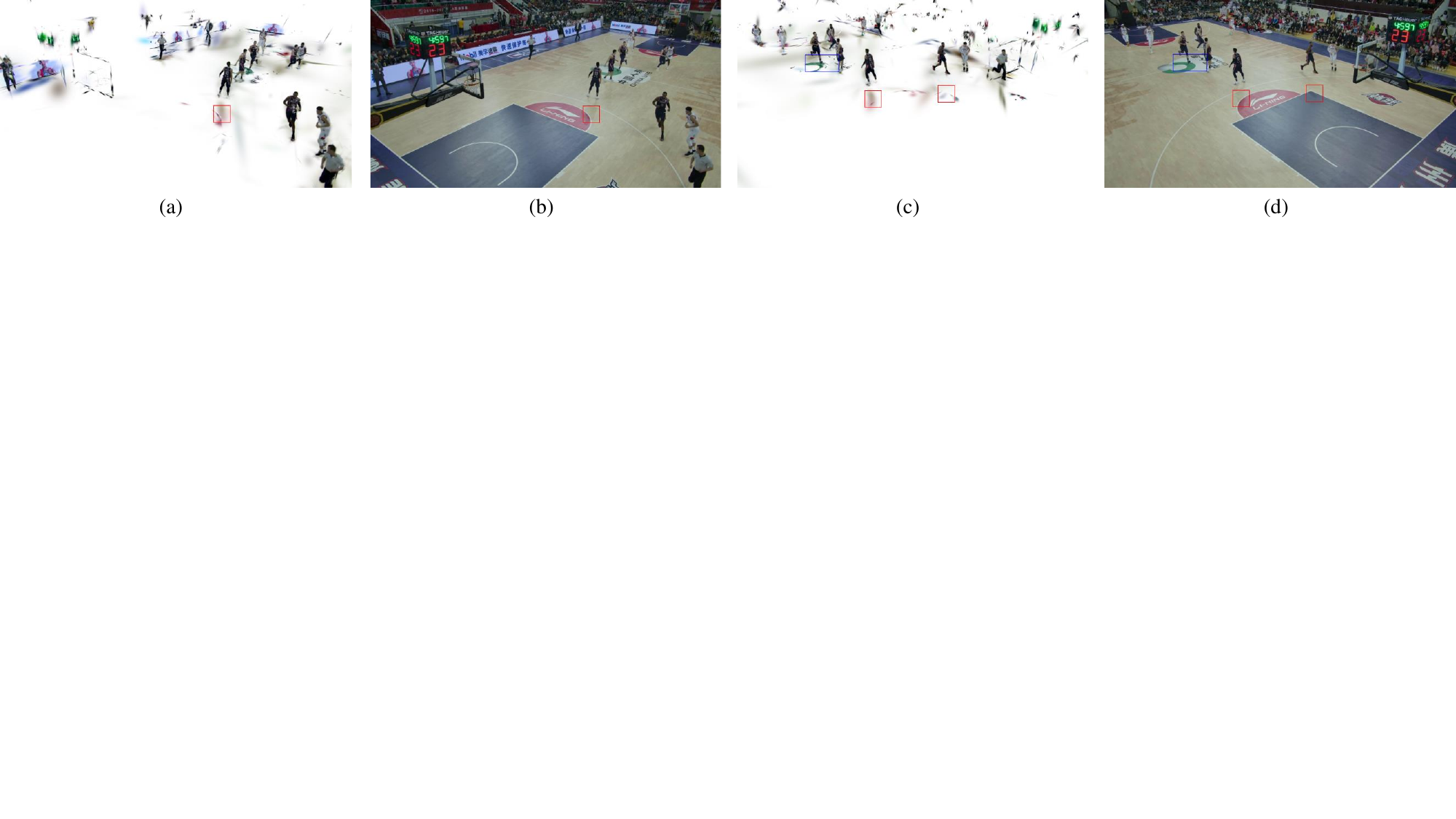}
\caption{\textbf{Basketball court dataset experiment.} (a) and (c) are  dynamic point renderings, while (b) and (d) are GT. The black floaters are actually Gaussian points from the dynamic background.}
\label{fig: basketball}
\vspace{-1em}
\end{figure}

Tab. \ref{tab: nv3d all quality comparison} and Fig. \ref{fig: nv3d comparison} respectively present the quantitative and qualitative evaluations of various methods on the N3DV video dataset. As shown in Tab. \ref{tab: nv3d all quality comparison}, our approach not only significantly surpasses previous methods in rendering quality but also achieves speeds at least 20 times faster compared to methods achieving similar rendering quality \cite{yang2023real,li2024spacetime}. It can be seen in Fig. \ref{fig: nv3d comparison} that our method not only achieves higher-quality modeling of static regions, such as the plate in the bottom right corner and the background outside the window in the \textit{coffee martini}, but also provides more detailed modeling of moving regions, such as the arms. As for the MeetRoom dataset results, shown in Fig. \ref{fig: meetroom compairl} and Tab. \ref{tab: meetroom all quality comparison}, our method achieves state-of-the-art performance in rendering quality, training time, and storage efficiency. Particularly noteworthy is the storage efficiency, as 3DGStream requires 30 times more storage compared to our approach. 
The results of training on the basketball court dataset are presented in Fig. \ref{fig: basketball} and the supplementary videos (basketball 1 and 2), showcasing our method's ability to handle highly complex dynamic scenes. Our decomposition technique effectively separates all athletes from the scene, illustrating the model's strong adaptability to occlusion, as discussed in Sec. \ref{sec:segmentation}.

\begin{wrapfigure}{r}{0 cm}
    \centering
    \includegraphics[width=0.5\textwidth, trim=0.2cm 8.8cm 16cm 0cm, clip]{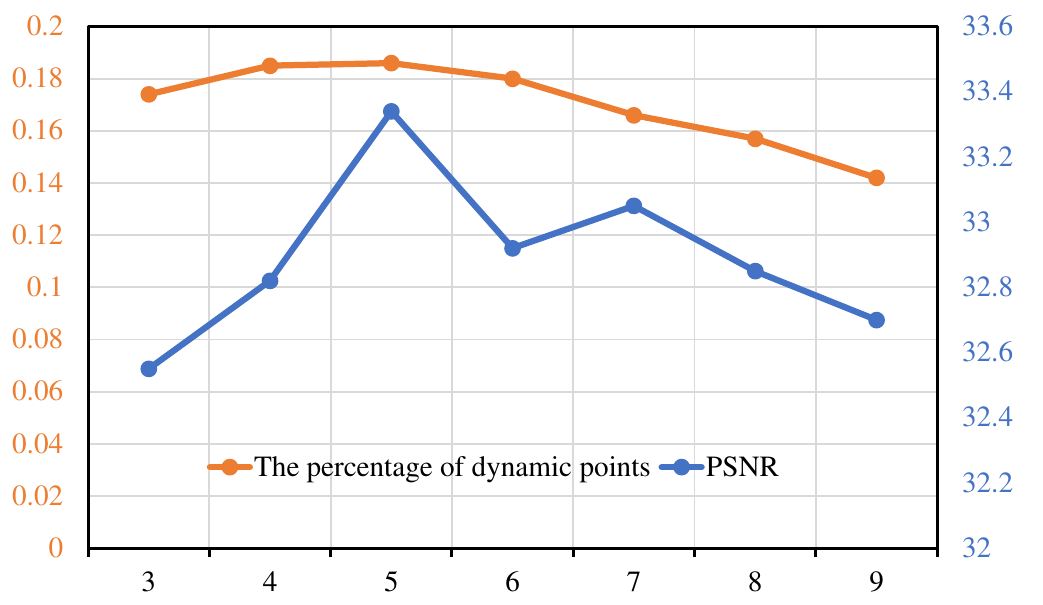}
    \caption{ \textbf{Distribution of dynamic points counts and PSNR at different thresholds.}}
    \label{fig:curve}
    \vspace{-1em}
\end{wrapfigure}

\subsection{ABLATION AND ANALYSIS}
\begin{figure}[t]
\centering
\vspace{-2em}
\setcounter{subfigure}{0}  
\subfloat[Lite]{\includegraphics[scale=0.071]{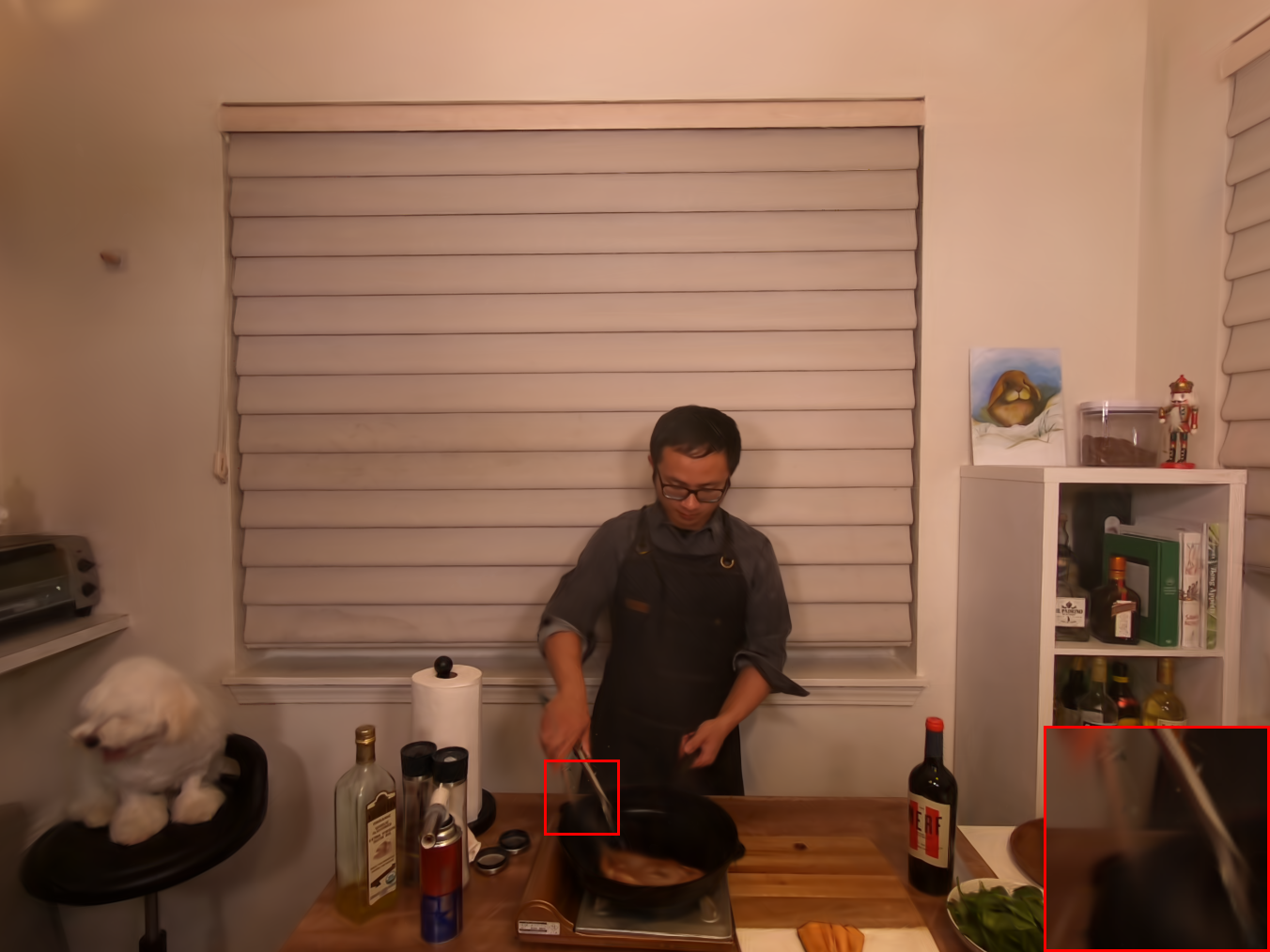}}
\hfill
\subfloat[Muti-plane encoder]{\includegraphics[scale=0.071]{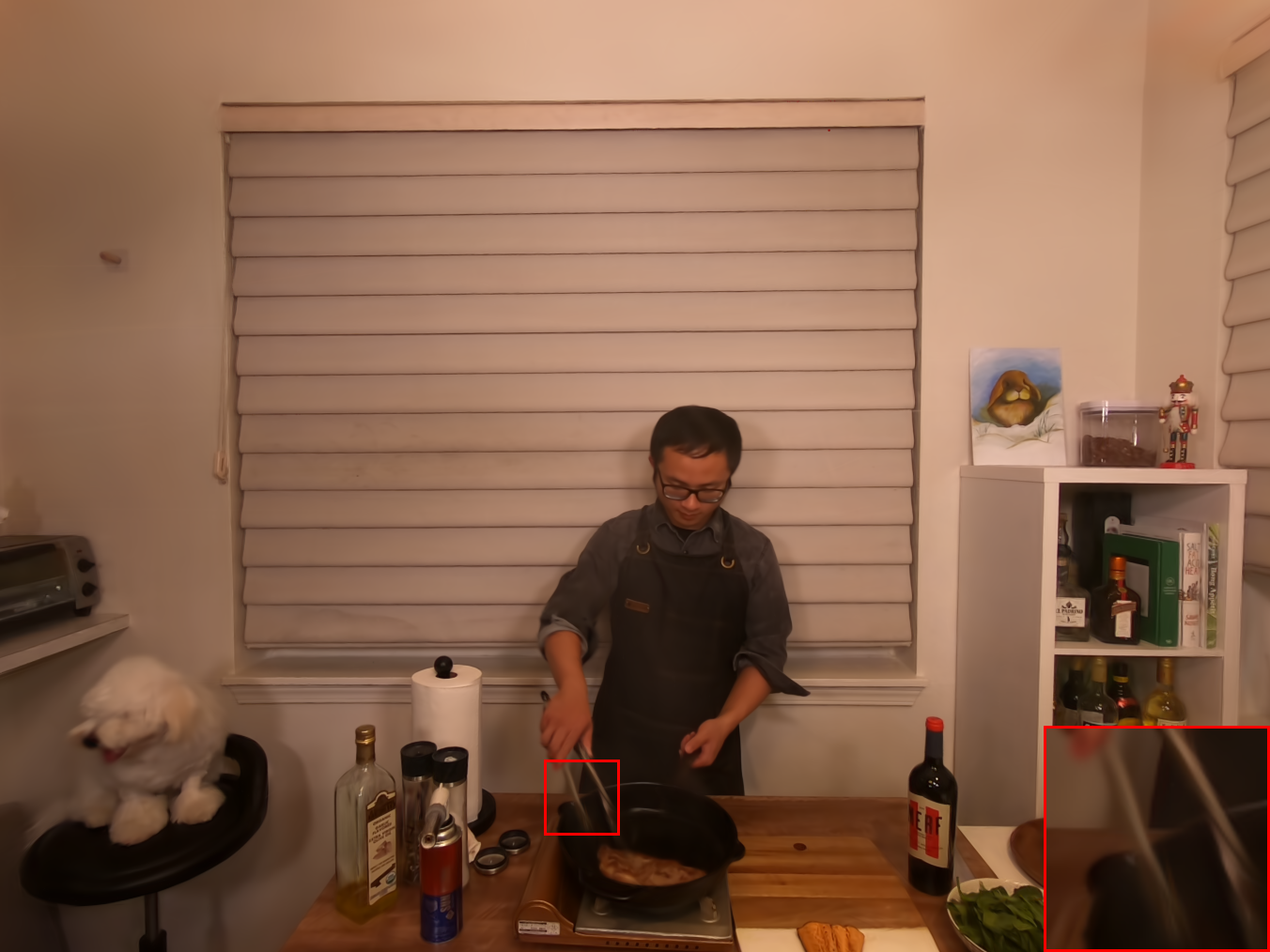}}
\hfill
\subfloat[W/O decomp]{\includegraphics[scale=0.071]{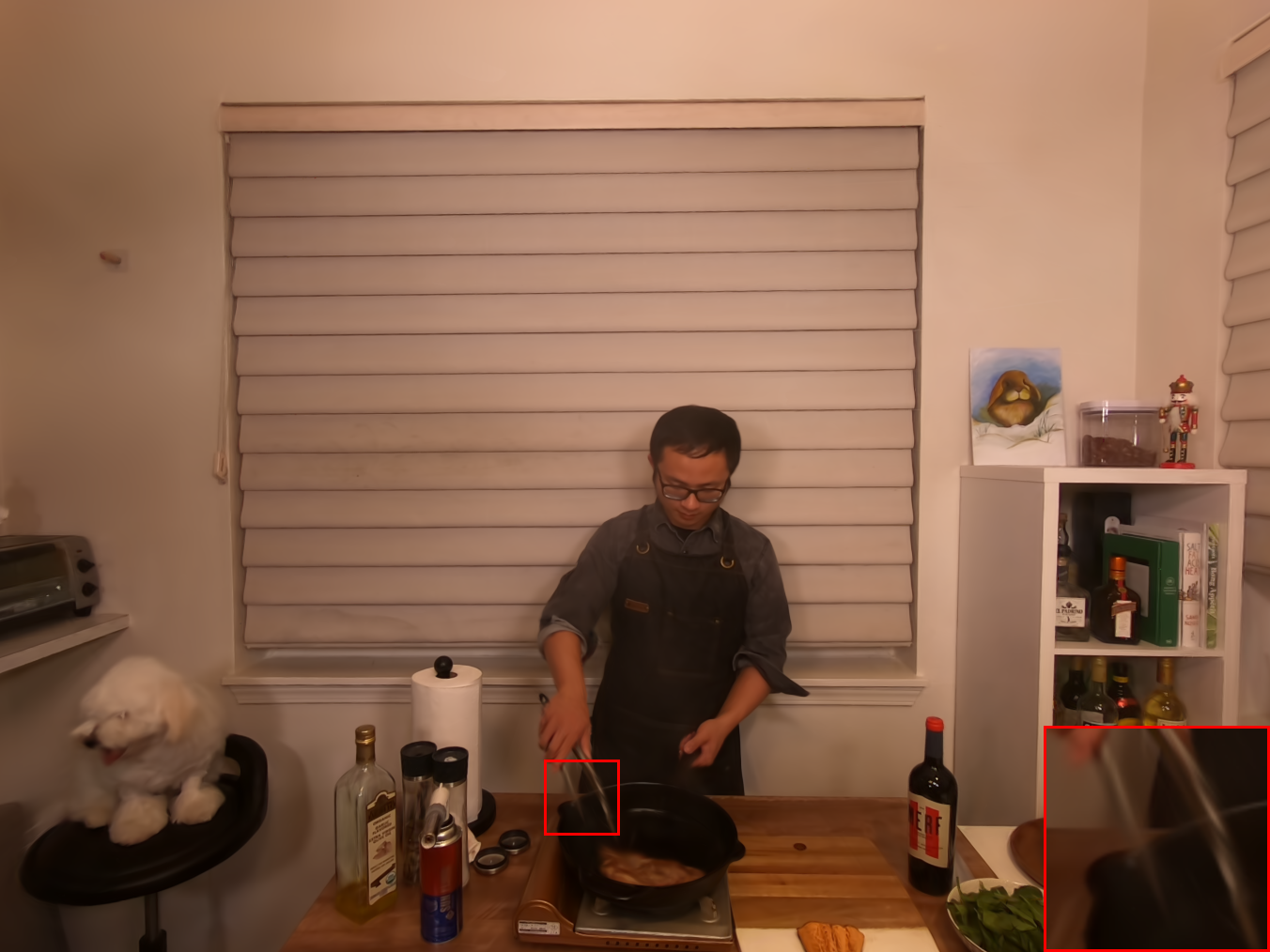}}
\hfill
\subfloat[Full(Ours)]{\includegraphics[scale=0.071]{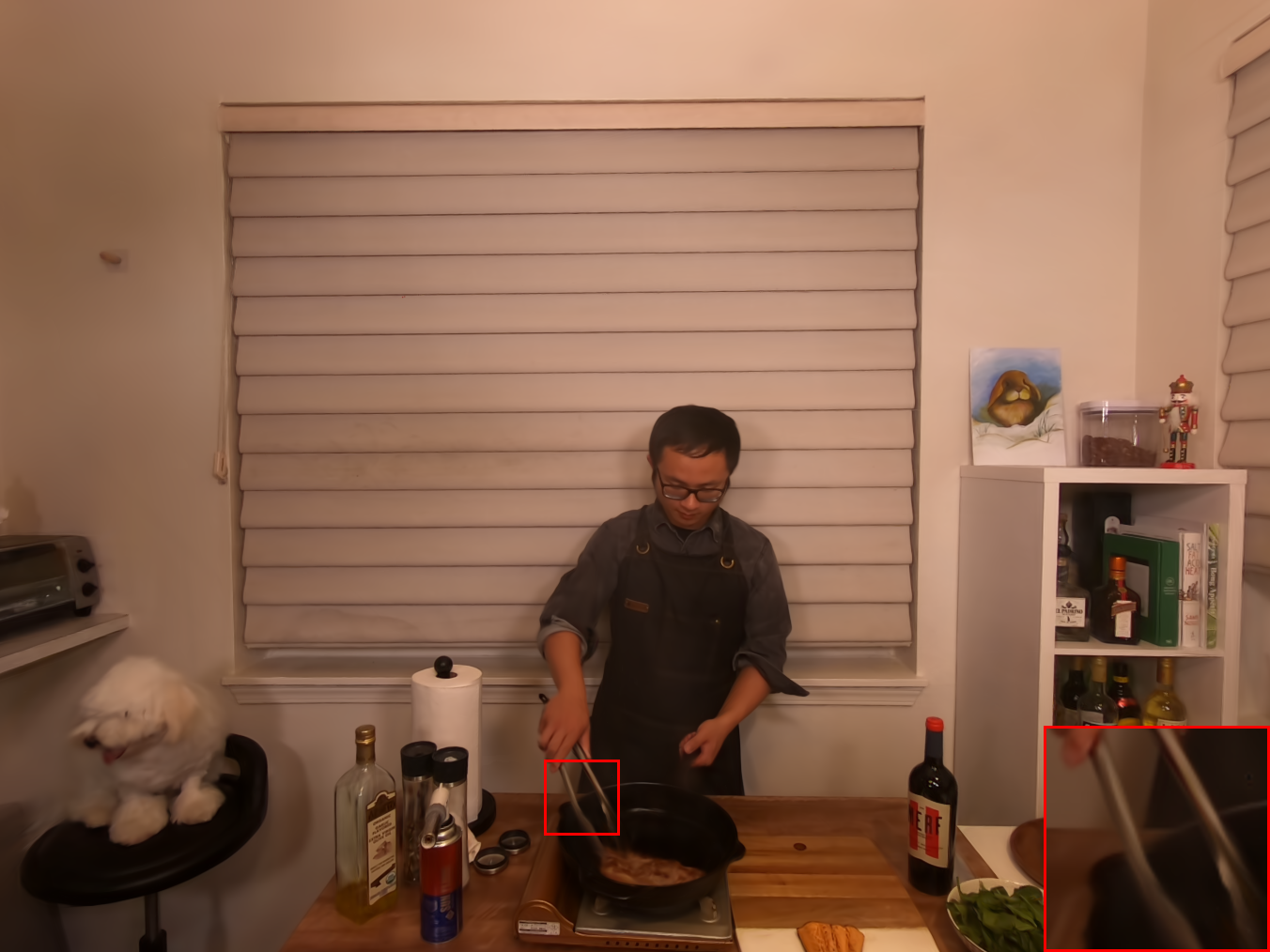}}
\setcounter{subfigure}{0}  


\caption{Some ablation experiments results on \textit{sear steak}.}
\label{fig: all ablation}
\end{figure}

\vspace{-1em}
\label{sec:ablation}
\textbf{Dynamic and static decomposition.} To validate the effectiveness of our dynamic-static decomposition method, we conducted experiments on the \textit{sear steak} and \textit{flame steak}. As illustrated in Fig. \ref{fig: all ablation}(c) and Tab. \ref{tab: ablation exp}, treating all points as dynamic led to increased computation time, significantly reduced rendering quality, and introduced blurring in areas with large motion amplitudes.

\textbf{Muti-plane encoder.} There are three commonly used choices for encoder selection: an implicit MLP \cite{gao2021dynamic}, multi-planes \cite{cao2023hexplane}, and a hash table \cite{muller2022instant}. In this study, we delve into employing the multi-plane approach to replace the 4D Hash as the encoder in our method. The subjective and objective experimental results, shown in Fig. \ref{fig: all ablation}(b) and Tab. \ref{tab: ablation exp}, indicate that while it slightly lags behind the hash table in terms of rendering speed and quality, it still outperforms methods that do not apply dynamic-static decomposition \cite{wu20244d}.

\begin{figure}[t]
    \centering
    \vspace{-1em}
    \includegraphics[width= \textwidth ,trim=0cm 13cm 0cm 2cm]{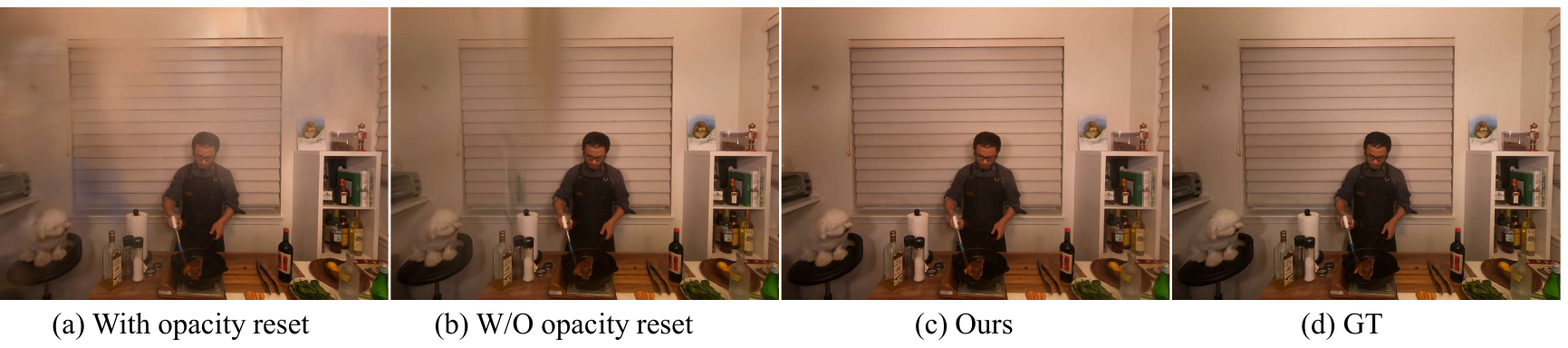}
\caption{\textbf{Importance Pruning Ablation Experiments:} (a), (b), (c), and (d) show the rendered results of the our model with opacity reset every 3000 iterations, without opacity reset, with our importance pruning method, and the ground truth, respectively.}
\label{fig: ablation prune}
\end{figure}

\textbf{Temporal importance pruning.} As shown in Fig. \ref{fig: ablation prune}, (a) and (b) exhibit severe artifacts. In contrast, images rendered with our pruning strategy, (c), appear much cleaner.

\section{Discussion}
\label{Sec:Discussion}

\textbf{Incomplete decomposition of dynamic and static points.} 
Although we employ the pixel level supervisor, it fails to fully decouple dynamic and static points. This can lead to two issues: Gaussian points in textureless regions of small moving objects may be mistaken for static, while static objects may be identified as dynamic due to interference from nearby moving objects.

In the first situation, as shown in Fig. \ref{figs:dyn and sta}(b), certain textureless areas, like clothing and the table, are mistakenly identified as static, despite being dynamic. In fact, this proves beneficial in Tab. \ref{tab: ablation exp}. If the entire table were labeled as dynamic, the increased dynamic points would lower rendering quality (W/o decomp). By recognizing only the edges as dynamic, where pixel changes are significant, the method reduces the number of dynamic points and enhances rendering quality (Ours).

In the second situation, as illustrated in Fig. \ref{fig: basketball}, some Gaussians in static areas are classified as dynamic due to the shadows or movements of the basketball players passing through these regions. Therefore, it is reasonable to identify these areas as dynamic. Classifying these points as static will impact the visual experience, as static points can hardly model dynamic areas.

\textbf{The selection of the dynamic threshold.} As shown in Fig. \ref{fig:curve}, experiments with the "cook spinach" scene revealed that varying the dynamic threshold \(\zeta\) from 3 to 9 did not significantly affect PSNR or the percentage of dynamic points, demonstrating the robustness of our method. To ensure consistency, we set the threshold to 7.

\textbf{The training time.} Due to the large dataset (nearly 6000 images), we load images during training, which empirically wastes around $40\%$ of the training time on disk I/O. Eliminating this overhead could reduce training time to 10 minutes while maintaining high-quality 4D scene reconstruction.

\section{Conclusion}
In this paper, we introduce \confDataName, which achieves fast convergence, compact storage, and high-quality real-time rendering capabilities within the field of 4D reconstruction.
The core innovation of our method lies in the introduction of a dynamic-static decomposition technique, which  can be applied to most existing dynamic scene reconstruction methods, enhancing quality and accelerating convergence. 
Additionally, we introduce a 4D Hash encoder and a multi-head decoder as our spatio-temporal structure, allowing for faster and more efficient temporal modeling of dynamic points. 
Finally, to prevent severe coupling between the canonical and deformation fields, we propose a novel temporal pruning method that effectively removes floaters in the scene. 
Our proposed method delivers competitive  results in just 5 minutes, and we hope it can offer new insights for applications struggling with training efficiency.


\textbf{Limitation:}  Similar to previous work \cite{sun20243dgstream,li2024spacetime}, our mrthod focuses on multi-view scenes and currently does not support monocular datasets for dynamic scene reconstruction. Additionally, our method focuses on scene reconstruction and does not include human reconstruction \cite{wu2020multi,cheng2023dna}.

\section{ACKNOWLEDGEMENTS}
This work is financially supported by Guangdong Provincial Key Laboratory of Ultra High Definition Immersive Media Technology(Grant No. 2024B1212010006), National Natural Science Foundation of China U21B2012, Shenzhen Science and Technology Program-Shenzhen Cultivation of Excellent Scientific and Technological Innovation Talents project(Grant No. RCJC20200714114435057), this work is also financially supported for Outstanding Talents Training Fund in Shenzhen.

\bibliography{iclr2025_conference}

\begin{thebibliography}{44}
\providecommand{\natexlab}[1]{#1}
\providecommand{\url}[1]{\texttt{#1}}
\expandafter\ifx\csname urlstyle\endcsname\relax
  \providecommand{\doi}[1]{doi: #1}\else
  \providecommand{\doi}{doi: \begingroup \urlstyle{rm}\Url}\fi

\bibitem[Barron et~al.(2021)Barron, Mildenhall, Tancik, Hedman, Martin-Brualla, and Srinivasan]{barron2021mip}
Jonathan~T Barron, Ben Mildenhall, Matthew Tancik, Peter Hedman, Ricardo Martin-Brualla, and Pratul~P Srinivasan.
\newblock Mip-nerf: A multiscale representation for anti-aliasing neural radiance fields.
\newblock In \emph{Proceedings of the IEEE/CVF international conference on computer vision}, pp.\  5855--5864, 2021.

\bibitem[Barron et~al.(2022)Barron, Mildenhall, Verbin, Srinivasan, and Hedman]{barron2022mip}
Jonathan~T Barron, Ben Mildenhall, Dor Verbin, Pratul~P Srinivasan, and Peter Hedman.
\newblock Mip-nerf 360: Unbounded anti-aliased neural radiance fields.
\newblock In \emph{Proceedings of the IEEE/CVF conference on computer vision and pattern recognition}, pp.\  5470--5479, 2022.

\bibitem[Barron et~al.(2023)Barron, Mildenhall, Verbin, Srinivasan, and Hedman]{barron2023zip}
Jonathan~T Barron, Ben Mildenhall, Dor Verbin, Pratul~P Srinivasan, and Peter Hedman.
\newblock Zip-nerf: Anti-aliased grid-based neural radiance fields.
\newblock In \emph{Proceedings of the IEEE/CVF International Conference on Computer Vision}, pp.\  19697--19705, 2023.

\bibitem[Cao \& Johnson(2023)Cao and Johnson]{cao2023hexplane}
Ang Cao and Justin Johnson.
\newblock Hexplane: A fast representation for dynamic scenes.
\newblock In \emph{Proceedings of the IEEE/CVF Conference on Computer Vision and Pattern Recognition}, pp.\  130--141, 2023.

\bibitem[Chen et~al.(2022)Chen, Xu, Geiger, Yu, and Su]{chen2022tensorf}
Anpei Chen, Zexiang Xu, Andreas Geiger, Jingyi Yu, and Hao Su.
\newblock Tensorf: Tensorial radiance fields.
\newblock In \emph{European conference on computer vision}, pp.\  333--350. Springer, 2022.

\bibitem[Cheng et~al.(2023)Cheng, Chen, Fan, Yin, Chen, Cai, Wang, Gao, Yu, Lin, et~al.]{cheng2023dna}
Wei Cheng, Ruixiang Chen, Siming Fan, Wanqi Yin, Keyu Chen, Zhongang Cai, Jingbo Wang, Yang Gao, Zhengming Yu, Zhengyu Lin, et~al.
\newblock Dna-rendering: A diverse neural actor repository for high-fidelity human-centric rendering.
\newblock In \emph{Proceedings of the IEEE/CVF International Conference on Computer Vision}, pp.\  19982--19993, 2023.

\bibitem[Deng et~al.(2024)Deng, Chen, Zhang, Yang, Yuan, Liu, Wang, Wang, and Chen]{deng2024compact}
Tianchen Deng, Yaohui Chen, Leyan Zhang, Jianfei Yang, Shenghai Yuan, Jiuming Liu, Danwei Wang, Hesheng Wang, and Weidong Chen.
\newblock Compact 3d gaussian splatting for dense visual slam.
\newblock \emph{arXiv preprint arXiv:2403.11247}, 2024.

\bibitem[Fan et~al.(2023)Fan, Wang, Wen, Zhu, Xu, and Wang]{fan2023lightgaussian}
Zhiwen Fan, Kevin Wang, Kairun Wen, Zehao Zhu, Dejia Xu, and Zhangyang Wang.
\newblock Lightgaussian: Unbounded 3d gaussian compression with 15x reduction and 200+ fps.
\newblock \emph{arXiv preprint arXiv:2311.17245}, 2023.

\bibitem[Fridovich-Keil et~al.(2022)Fridovich-Keil, Yu, Tancik, Chen, Recht, and Kanazawa]{fridovich2022plenoxels}
Sara Fridovich-Keil, Alex Yu, Matthew Tancik, Qinhong Chen, Benjamin Recht, and Angjoo Kanazawa.
\newblock Plenoxels: Radiance fields without neural networks.
\newblock In \emph{Proceedings of the IEEE/CVF conference on computer vision and pattern recognition}, pp.\  5501--5510, 2022.

\bibitem[Fridovich-Keil et~al.(2023)Fridovich-Keil, Meanti, Warburg, Recht, and Kanazawa]{fridovich2023k}
Sara Fridovich-Keil, Giacomo Meanti, Frederik~Rahb{\ae}k Warburg, Benjamin Recht, and Angjoo Kanazawa.
\newblock K-planes: Explicit radiance fields in space, time, and appearance.
\newblock In \emph{Proceedings of the IEEE/CVF Conference on Computer Vision and Pattern Recognition}, pp.\  12479--12488, 2023.

\bibitem[Gao et~al.(2021)Gao, Saraf, Kopf, and Huang]{gao2021dynamic}
Chen Gao, Ayush Saraf, Johannes Kopf, and Jia-Bin Huang.
\newblock Dynamic view synthesis from dynamic monocular video.
\newblock In \emph{Proceedings of the IEEE/CVF International Conference on Computer Vision}, pp.\  5712--5721, 2021.

\bibitem[He et~al.(2024)He, Chen, Lu, Song, and Zhang]{he2024s4d}
Bing He, Yunuo Chen, Guo Lu, Li~Song, and Wenjun Zhang.
\newblock S4d: Streaming 4d real-world reconstruction with gaussians and 3d control points.
\newblock \emph{arXiv preprint arXiv:2408.13036}, 2024.

\bibitem[Huang et~al.(2024{\natexlab{a}})Huang, Yu, Chen, Geiger, and Gao]{huang20242d}
Binbin Huang, Zehao Yu, Anpei Chen, Andreas Geiger, and Shenghua Gao.
\newblock 2d gaussian splatting for geometrically accurate radiance fields.
\newblock \emph{arXiv preprint arXiv:2403.17888}, 2024{\natexlab{a}}.

\bibitem[Huang et~al.(2024{\natexlab{b}})Huang, Sun, Yang, Lyu, Cao, and Qi]{huang2024sc}
Yi-Hua Huang, Yang-Tian Sun, Ziyi Yang, Xiaoyang Lyu, Yan-Pei Cao, and Xiaojuan Qi.
\newblock Sc-gs: Sparse-controlled gaussian splatting for editable dynamic scenes.
\newblock In \emph{Proceedings of the IEEE/CVF Conference on Computer Vision and Pattern Recognition}, pp.\  4220--4230, 2024{\natexlab{b}}.

\bibitem[Kerbl et~al.(2023)Kerbl, Kopanas, Leimk{\"u}hler, and Drettakis]{kerbl20233d}
Bernhard Kerbl, Georgios Kopanas, Thomas Leimk{\"u}hler, and George Drettakis.
\newblock 3d gaussian splatting for real-time radiance field rendering.
\newblock \emph{ACM Trans. Graph.}, 42\penalty0 (4):\penalty0 139--1, 2023.

\bibitem[Kingma \& Ba(2014)Kingma and Ba]{kingma2014adam}
Diederik~P Kingma and Jimmy Ba.
\newblock Adam: A method for stochastic optimization.
\newblock \emph{arXiv preprint arXiv:1412.6980}, 2014.

\bibitem[Kratimenos et~al.(2023)Kratimenos, Lei, and Daniilidis]{kratimenos2023dynmf}
Agelos Kratimenos, Jiahui Lei, and Kostas Daniilidis.
\newblock Dynmf: Neural motion factorization for real-time dynamic view synthesis with 3d gaussian splatting.
\newblock \emph{arXiv preprint arXiv:2312.00112}, 2023.

\bibitem[Li et~al.(2022{\natexlab{a}})Li, Shen, Wang, Shen, and Tan]{li2022streaming}
Lingzhi Li, Zhen Shen, Zhongshu Wang, Li~Shen, and Ping Tan.
\newblock Streaming radiance fields for 3d video synthesis.
\newblock \emph{Advances in Neural Information Processing Systems}, 35:\penalty0 13485--13498, 2022{\natexlab{a}}.

\bibitem[Li et~al.(2022{\natexlab{b}})Li, Slavcheva, Zollhoefer, Green, Lassner, Kim, Schmidt, Lovegrove, Goesele, Newcombe, et~al.]{li2022neural}
Tianye Li, Mira Slavcheva, Michael Zollhoefer, Simon Green, Christoph Lassner, Changil Kim, Tanner Schmidt, Steven Lovegrove, Michael Goesele, Richard Newcombe, et~al.
\newblock Neural 3d video synthesis from multi-view video.
\newblock In \emph{Proceedings of the IEEE/CVF Conference on Computer Vision and Pattern Recognition}, pp.\  5521--5531, 2022{\natexlab{b}}.

\bibitem[Li et~al.(2024)Li, Chen, Li, and Xu]{li2024spacetime}
Zhan Li, Zhang Chen, Zhong Li, and Yi~Xu.
\newblock Spacetime gaussian feature splatting for real-time dynamic view synthesis.
\newblock In \emph{Proceedings of the IEEE/CVF Conference on Computer Vision and Pattern Recognition}, pp.\  8508--8520, 2024.

\bibitem[Li et~al.(2021)Li, Niklaus, Snavely, and Wang]{li2021neural}
Zhengqi Li, Simon Niklaus, Noah Snavely, and Oliver Wang.
\newblock Neural scene flow fields for space-time view synthesis of dynamic scenes.
\newblock In \emph{Proceedings of the IEEE/CVF Conference on Computer Vision and Pattern Recognition}, pp.\  6498--6508, 2021.

\bibitem[Liang et~al.(2023)Liang, Khan, Li, Nguyen-Phuoc, Lanman, Tompkin, and Xiao]{liang2023gaufre}
Yiqing Liang, Numair Khan, Zhengqin Li, Thu Nguyen-Phuoc, Douglas Lanman, James Tompkin, and Lei Xiao.
\newblock Gaufre: Gaussian deformation fields for real-time dynamic novel view synthesis.
\newblock \emph{arXiv preprint arXiv:2312.11458}, 2023.

\bibitem[Lin et~al.(2024)Lin, Dai, Zhu, and Yao]{lin2024gaussian}
Youtian Lin, Zuozhuo Dai, Siyu Zhu, and Yao Yao.
\newblock Gaussian-flow: 4d reconstruction with dynamic 3d gaussian particle.
\newblock In \emph{Proceedings of the IEEE/CVF Conference on Computer Vision and Pattern Recognition}, pp.\  21136--21145, 2024.

\bibitem[Lu et~al.(2024)Lu, Yu, Xu, Xiangli, Wang, Lin, and Dai]{lu2024scaffold}
Tao Lu, Mulin Yu, Linning Xu, Yuanbo Xiangli, Limin Wang, Dahua Lin, and Bo~Dai.
\newblock Scaffold-gs: Structured 3d gaussians for view-adaptive rendering.
\newblock In \emph{Proceedings of the IEEE/CVF Conference on Computer Vision and Pattern Recognition}, pp.\  20654--20664, 2024.

\bibitem[Mildenhall et~al.(2021)Mildenhall, Srinivasan, Tancik, Barron, Ramamoorthi, and Ng]{mildenhall2021nerf}
Ben Mildenhall, Pratul~P Srinivasan, Matthew Tancik, Jonathan~T Barron, Ravi Ramamoorthi, and Ren Ng.
\newblock Nerf: Representing scenes as neural radiance fields for view synthesis.
\newblock \emph{Communications of the ACM}, 65\penalty0 (1):\penalty0 99--106, 2021.

\bibitem[M{\"u}ller et~al.(2022)M{\"u}ller, Evans, Schied, and Keller]{muller2022instant}
Thomas M{\"u}ller, Alex Evans, Christoph Schied, and Alexander Keller.
\newblock Instant neural graphics primitives with a multiresolution hash encoding.
\newblock \emph{ACM transactions on graphics (TOG)}, 41\penalty0 (4):\penalty0 1--15, 2022.

\bibitem[Niemeyer et~al.(2024)Niemeyer, Manhardt, Rakotosaona, Oechsle, Duckworth, Gosula, Tateno, Bates, Kaeser, and Tombari]{niemeyer2024radsplat}
Michael Niemeyer, Fabian Manhardt, Marie-Julie Rakotosaona, Michael Oechsle, Daniel Duckworth, Rama Gosula, Keisuke Tateno, John Bates, Dominik Kaeser, and Federico Tombari.
\newblock Radsplat: Radiance field-informed gaussian splatting for robust real-time rendering with 900+ fps.
\newblock \emph{arXiv preprint arXiv:2403.13806}, 2024.

\bibitem[Park et~al.(2021)Park, Sinha, Barron, Bouaziz, Goldman, Seitz, and Martin-Brualla]{park2021nerfies}
Keunhong Park, Utkarsh Sinha, Jonathan~T Barron, Sofien Bouaziz, Dan~B Goldman, Steven~M Seitz, and Ricardo Martin-Brualla.
\newblock Nerfies: Deformable neural radiance fields.
\newblock In \emph{Proceedings of the IEEE/CVF International Conference on Computer Vision}, pp.\  5865--5874, 2021.

\bibitem[Pumarola et~al.(2021)Pumarola, Corona, Pons-Moll, and Moreno-Noguer]{pumarola2021d}
Albert Pumarola, Enric Corona, Gerard Pons-Moll, and Francesc Moreno-Noguer.
\newblock D-nerf: Neural radiance fields for dynamic scenes.
\newblock In \emph{Proceedings of the IEEE/CVF Conference on Computer Vision and Pattern Recognition}, pp.\  10318--10327, 2021.

\bibitem[Reiser et~al.(2023)Reiser, Szeliski, Verbin, Srinivasan, Mildenhall, Geiger, Barron, and Hedman]{reiser2023merf}
Christian Reiser, Rick Szeliski, Dor Verbin, Pratul Srinivasan, Ben Mildenhall, Andreas Geiger, Jon Barron, and Peter Hedman.
\newblock Merf: Memory-efficient radiance fields for real-time view synthesis in unbounded scenes.
\newblock \emph{ACM Transactions on Graphics (TOG)}, 42\penalty0 (4):\penalty0 1--12, 2023.

\bibitem[Schonberger \& Frahm(2016)Schonberger and Frahm]{schonberger2016structure}
Johannes~L Schonberger and Jan-Michael Frahm.
\newblock Structure-from-motion revisited.
\newblock In \emph{Proceedings of the IEEE conference on computer vision and pattern recognition}, pp.\  4104--4113, 2016.

\bibitem[Song et~al.(2023)Song, Chen, Li, Chen, Chen, Yuan, Xu, and Geiger]{song2023nerfplayer}
Liangchen Song, Anpei Chen, Zhong Li, Zhang Chen, Lele Chen, Junsong Yuan, Yi~Xu, and Andreas Geiger.
\newblock Nerfplayer: A streamable dynamic scene representation with decomposed neural radiance fields.
\newblock \emph{IEEE Transactions on Visualization and Computer Graphics}, 29\penalty0 (5):\penalty0 2732--2742, 2023.

\bibitem[Sun et~al.(2022)Sun, Sun, and Chen]{sun2022direct}
Cheng Sun, Min Sun, and Hwann-Tzong Chen.
\newblock Direct voxel grid optimization: Super-fast convergence for radiance fields reconstruction.
\newblock In \emph{Proceedings of the IEEE/CVF conference on computer vision and pattern recognition}, pp.\  5459--5469, 2022.

\bibitem[Sun et~al.(2024)Sun, Jiao, Li, Zhang, Zhao, and Xing]{sun20243dgstream}
Jiakai Sun, Han Jiao, Guangyuan Li, Zhanjie Zhang, Lei Zhao, and Wei Xing.
\newblock 3dgstream: On-the-fly training of 3d gaussians for efficient streaming of photo-realistic free-viewpoint videos.
\newblock In \emph{Proceedings of the IEEE/CVF Conference on Computer Vision and Pattern Recognition}, pp.\  20675--20685, 2024.

\bibitem[VRU(2024)]{VRU}
VRU.
\newblock https://anonymous.4open.science/r/vru-sequence/.
\newblock 2024.

\bibitem[Wang et~al.(2023)Wang, Tan, Li, Tian, Song, and Liu]{wang2023mixed}
Feng Wang, Sinan Tan, Xinghang Li, Zeyue Tian, Yafei Song, and Huaping Liu.
\newblock Mixed neural voxels for fast multi-view video synthesis.
\newblock In \emph{Proceedings of the IEEE/CVF International Conference on Computer Vision}, pp.\  19706--19716, 2023.

\bibitem[Wu et~al.(2024)Wu, Yi, Fang, Xie, Zhang, Wei, Liu, Tian, and Wang]{wu20244d}
Guanjun Wu, Taoran Yi, Jiemin Fang, Lingxi Xie, Xiaopeng Zhang, Wei Wei, Wenyu Liu, Qi~Tian, and Xinggang Wang.
\newblock 4d gaussian splatting for real-time dynamic scene rendering.
\newblock In \emph{Proceedings of the IEEE/CVF Conference on Computer Vision and Pattern Recognition}, pp.\  20310--20320, 2024.

\bibitem[Wu et~al.(2020)Wu, Wang, Hu, and Yu]{wu2020multi}
Minye Wu, Yuehao Wang, Qiang Hu, and Jingyi Yu.
\newblock Multi-view neural human rendering.
\newblock In \emph{Proceedings of the IEEE/CVF Conference on Computer Vision and Pattern Recognition}, pp.\  1682--1691, 2020.

\bibitem[Yan et~al.()Yan, Peng, Tang, and Wang]{yan20244d}
Jinbo Yan, Rui Peng, Luyang Tang, and Ronggang Wang.
\newblock 4d gaussian splatting with scale-aware residual field and adaptive optimization for real-time rendering of temporally complex dynamic scenes.
\newblock In \emph{ACM Multimedia 2024}.

\bibitem[Yang et~al.(2023)Yang, Yang, Pan, Zhu, and Zhang]{yang2023real}
Zeyu Yang, Hongye Yang, Zijie Pan, Xiatian Zhu, and Li~Zhang.
\newblock Real-time photorealistic dynamic scene representation and rendering with 4d gaussian splatting.
\newblock \emph{arXiv preprint arXiv:2310.10642}, 2023.

\bibitem[Yang et~al.(2024)Yang, Gao, Zhou, Jiao, Zhang, and Jin]{yang2024deformable}
Ziyi Yang, Xinyu Gao, Wen Zhou, Shaohui Jiao, Yuqing Zhang, and Xiaogang Jin.
\newblock Deformable 3d gaussians for high-fidelity monocular dynamic scene reconstruction.
\newblock In \emph{Proceedings of the IEEE/CVF Conference on Computer Vision and Pattern Recognition}, pp.\  20331--20341, 2024.

\bibitem[Yu et~al.(2021)Yu, Li, Tancik, Li, Ng, and Kanazawa]{yu2021plenoctrees}
Alex Yu, Ruilong Li, Matthew Tancik, Hao Li, Ren Ng, and Angjoo Kanazawa.
\newblock Plenoctrees for real-time rendering of neural radiance fields.
\newblock In \emph{Proceedings of the IEEE/CVF International Conference on Computer Vision}, pp.\  5752--5761, 2021.

\bibitem[Yu et~al.(2024)Yu, Chen, Huang, Sattler, and Geiger]{yu2024mip}
Zehao Yu, Anpei Chen, Binbin Huang, Torsten Sattler, and Andreas Geiger.
\newblock Mip-splatting: Alias-free 3d gaussian splatting.
\newblock In \emph{Proceedings of the IEEE/CVF Conference on Computer Vision and Pattern Recognition}, pp.\  19447--19456, 2024.

\bibitem[Zhang et~al.(2018)Zhang, Isola, Efros, Shechtman, and Wang]{zhang2018unreasonable}
Richard Zhang, Phillip Isola, Alexei~A Efros, Eli Shechtman, and Oliver Wang.
\newblock The unreasonable effectiveness of deep features as a perceptual metric.
\newblock In \emph{Proceedings of the IEEE conference on computer vision and pattern recognition}, pp.\  586--595, 2018.

\end{thebibliography}
\bibliographystyle{iclr2025_conference}



\end{document}